\documentclass[journal]{IEEEtran}

\usepackage[T1]{fontenc}
\usepackage[hyphens]{url}
\usepackage{graphicx}
\usepackage{amsmath}
\usepackage{amssymb}
\usepackage{amsfonts}
\usepackage{mathrsfs}
\usepackage{textcomp}
\usepackage{booktabs}
\usepackage{bm}
\usepackage{pifont}
\usepackage{cite}
\usepackage{xcolor}
\definecolor{citecolor}{HTML}{0071bc}
\usepackage[colorlinks, linkcolor=red,  anchorcolor=blue, citecolor=citecolor]{hyperref} 

\usepackage{xcolor}
\definecolor{SeaGreen4}{RGB}{0,205,102} 
\definecolor{SlateBlue}{RGB}{106,90,205} 
\definecolor{DarkRed}{RGB}{178,34,34} 
	
\title{Physics-Informed Neural Networks for Complex Eigenfrequency Identification and Mode Structure Reconstruction of the Ground-State ITG Branch}
\author{Dengdi Sun, Bingbing Zhang, Xiao Wang, Zikang Yan, Yuqiang Tao*, Qingquan Yang*, \\ 
    Guosheng Xu, and Jin Tang
\thanks{$\bullet$  Dengdi Sun, Bingbing Zhang are with School of Artificial Intelligence, Anhui University, Hefei, China. (email: sundengdi@ahu.edu.cn, 3280689048@qq.com)} 
\thanks{$\bullet$ Xiao Wang, Zikang Yan, Jin Tang are with School of Computer Science and Technology, Anhui University, Hefei, China. (email: \{xiaowang, tangjin\}@ahu.edu.cn, e24201056@stu.ahu.edu.cn)}  
\thanks{$\bullet$ Yuqiang Tao is with School of Physics and Electronic Information, Anhui Normal University, Wuhu, China. (email: yuqiang.tao@ahnu.edu.cn)}  
\thanks{$\bullet$ Qingquan Yang, Guosheng Xu are with Institute of Plasma Physics, Chinese Academy of Sciences, Hefei, China. (email: yangqq@ipp.ac.cn, gsxu@ipp.ac.cn)}
\thanks{* Corresponding Author: Yuqiang Tao $\&$ Qingquan Yang}
    }

\begin{document}

\maketitle

\begin{abstract}
Physics-informed neural networks (PINNs) combine sparse observations with physical equations, providing an important approach for modeling complex plasma processes and inferring unknown physical quantities. The steep-gradient pedestal of high-confinement-mode tokamaks is closely linked to plasma confinement and edge transport. Analyzing ion-temperature-gradient (ITG) drift waves in this region requires jointly identifying complex eigenfrequencies and reconstructing two-dimensional complex-valued mode fields. Localized high-frequency oscillations, strong real-imaginary coupling, and nonlinear coupling between the mode field and eigenfrequency challenge PINN representation and joint optimization. To address these challenges, we propose a physics-informed neural framework combining Fourier feature encoding, complex-valued feature propagation, and three-stage training. Under sparse observations and physical constraints, it jointly solves for the complex eigenfrequency and mode field of a representative ground-state ITG branch. Experiments show that the framework accurately recovers the target complex eigenfrequency and two-dimensional complex-valued mode field and outperforms representative PINN baselines. It also provides a basis for analyzing higher-order and multiple-branch drift-wave modes. 
\end{abstract}


\section{Introduction}

\IEEEPARstart{P}{hysics-informed} inverse learning has become an important paradigm for recovering hidden states and unknown physical parameters from sparse observations~\cite{raissi2019physics, karniadakis2021physics, raissi2020hidden, yang2021bpinns, yang2022multioutput}. In many complex physical systems, such as magnetically confined plasmas, scientific inverse problems require not only the reconstruction of physical fields but also the simultaneous identification of key physical parameters~\cite{jin2022quantum, holliday2023solving}. Because these unknown physical parameters are tightly coupled with the physical fields through the governing equations, sparse observations can only provide partial information about the target field, thereby leading to serious problems such as unstable joint optimization and difficulty in representing oscillatory structures. This poses a great challenge to standard physics-informed neural networks (PINNs). The simulation of ion-temperature-gradient (ITG) mode instabilities in fusion plasmas is exactly such a typical complex inverse problem. 

\begin{figure}[t!]
    \centering
    \includegraphics[width=\columnwidth]{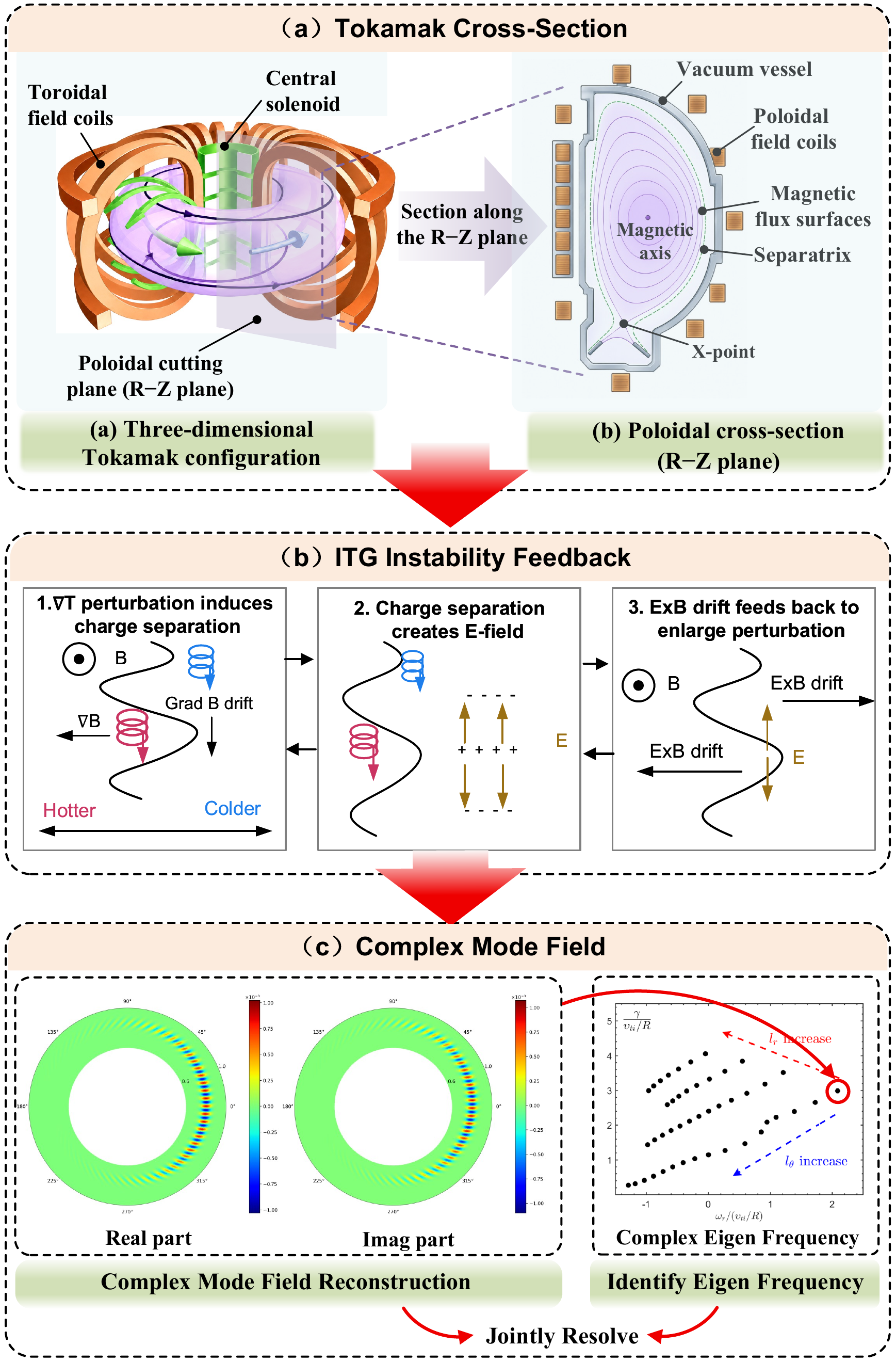}
    \caption{ Illustration of ITG instability feedback mechanism inverse problem resolution. (a) The tokamak configuration and the poloidal cutting plane. (b) ITG feedback process occurs at the plasma boundary during the H mode, involving differential magnetic drifts, charge separation, electrostatic-field generation, and $\mathbf{E}\times\mathbf{B}$ drift amplification. (c) The ITG process is simulated physically using a two-dimensional complex mode field, and the solution of the inverse problem requires simultaneously reconstructing the real and imaginary parts of the complex mode field in the annular computational domain and identifying its eigenfrequency.}
    \label{fig:tokamak3d}
\end{figure}

As shown in Figure~\ref{fig:tokamak3d}, ITG modes are gradient-driven drift-wave microinstabilities associated with ion-scale turbulence and anomalous transport~\cite{Frei2022CollisionalITG, Wang2024ITGBarrier, leppin2023complex}. They typically arise in steep-gradient regions near the outer low-field side, where temperature-gradient-driven perturbations produce charge separation and an electric-field-mediated $E\times B$ feedback loop~\cite{Abdoul2015GlobalITG}. For a given ITG branch, its instability characteristics are not only described by a complex eigenfrequency, but also correspond to the two-dimensional complex mode field defined over the radial position and poloidal angle~\cite{Abdoul2017Ballooning2D, Qiu2024TwoDimensionalITG}. The real and imaginary parts of this field typically exhibit localized, high-frequency oscillatory stripe-like structures and are concentrated in specific poloidal regions. Therefore, the inverse problem of this physical process is to simultaneously identify the complex eigenfrequency of the target ITG branch and recover its corresponding 2D complex mode field under sparse observations of the mode field.

Traditionally, this problem usually relies on well-specified computational settings, such as governing operators and boundary conditions, and employs numerical discretization methods to solve the global eigenvalue problem for computing the eigenfrequency and mode structure~\cite{xie2016global,tao2025numerical}. In contrast, in inverse-problem scenarios where only sparse observations of the mode field are available, traditional numerical methods are difficult to use directly with limited observations for inversion. They also suffer from high computational cost and low computational speed, which are further exacerbated under steep-gradient conditions. Recently, PINNs provide a natural framework by combining observations with governing-equation and boundary constraints~\cite{raissi2019physics,karniadakis2021physics}; however, standard PINNs remain difficult to apply to this high-frequency oscillating, tightly coupled complex inverse problem.



The aforementioned difficulties mainly arise from \textbf{two key challenges:} \textbf{\ding{182} Modeling complexity:} Under steep-gradient conditions, the ITG mode field usually exhibits localized, high-frequency oscillatory spatial structures~\cite{wang2022when, krishnapriyan2021failure, rathore2024challenges, tancik2020fourier}. In addition, the mode field and the eigenfrequency are both complex-valued. Their real and imaginary parts are closely coupled and cannot be modeled as independent variables~\cite{trabelsi2018deep}. \textbf{\ding{183} Optimization Difficulty:} The complex eigenfrequency and the mode field are tightly coupled through the governing equation, and jointly optimizing them is clearly ill-conditioned. This makes the training process highly sensitive to initialization, loss balancing, and residual scaling~\cite{rathore2024challenges}.


Motivated by these challenges, we propose a novel \textbf{PINN} framework for \textbf{C}omplex \textbf{E}igenfrequency \textbf{I}dentification and 2D complex mode-field reconstruction of the ground-state ITG branch under sparse supervision (\textbf{CEI-PINN}). The framework parameterizes the complex mode field with a neural network, treats the complex eigenfrequency as an explicit trainable parameter, and enforces the full-domain complex eigenvalue equation as the physical constraint. Specifically, \textbf{for challenge \ding{182}}, we propose a complex mode-field representation with Fourier feature encoding and complex feature propagation. \textbf{For challenge \ding{183}}, we design a three-stage training strategy based on a Physics-informed learning objective. With approximately 6\% sparse supervision, the framework recovers a complex eigenfrequency close to the reference value and reconstructs the corresponding 2D complex mode field, demonstrating the feasibility of physics-informed complex eigenfrequency inversion from sparse observations. 
The main contributions of this paper are summarized as follows:

1) We formulate ITG complex eigenfrequency identification and 2D complex mode-field reconstruction under sparse supervision as a physics-informed complex eigenfrequency inverse problem, and propose a complex PINN framework for jointly recovering the mode field and its corresponding complex eigenfrequency.

2) We design a frequency-aware complex mode-field representation module that combines Fourier feature encoding with complex feature propagation, which enhances representation for oscillatory and high-frequency components while accurately characterizing the coupling between real and imaginary parts of the complex mode field.

3) We propose a three-stage training strategy that separates mode-field initialization, complex eigenfrequency identification, and mode-field refinement. After the complex eigenfrequency is identified, it is fixed for reconstruction, alleviating parameter compensation and joint-optimization instability between the mode field and the complex eigenfrequency.

\section{Related Work} 

\subsection{Steep-Gradient Drift-Wave Eigenanalysis}
Drift-wave instabilities have been extensively investigated through theoretical and numerical studies. Under conventional or non-steep-gradient conditions, local ballooning modes are commonly used to describe modes localized near the outboard midplane~\cite{du2017properties,chen2018strong,connor1978shear,singh2014finite}. Traditional approaches discretize the governing operator into a matrix eigenvalue problem ~\cite{saad2011eigenvalue,lehoucq1998arpack}. Under steep-gradient conditions, however, multiple eigenfrequency branches with two-dimensional structures may emerge, while electromagnetic and short-wavelength effects may require a global two-dimensional model ~\cite{xie2015unconventional,han2017multiple,zocco2018threshold}. In contrast to previous studies that primarily perform forward eigenvalue calculations based on complete physical models, our work recovers the complex eigenfrequency and its corresponding complex-valued mode field from sparse observations under physics-based constraints, providing a data- and physics-integrated approach to eigenmode analysis when observational information is incomplete.

\subsection{Physics-Informed Eigenvalue Inversion}
Physics-informed neural networks combine governing-equation residuals, boundary conditions, and sparse observations to infer unknown fields and physical parameters~\cite{raissi2019physics,karniadakis2021physics,cuomo2022scientific}. They have been widely applied to PDE-constrained inverse problems involving field reconstruction and parameter identification~\cite{wang2021gradient,wang2022when,rathore2024challenges}. In plasma and fusion research, physics-informed learning has also been used for transport modeling and equilibrium reconstruction~\cite{seo2024transport,bonotto2024planet,jang2024grad}. Unlike these settings, a complex eigenfrequency and a two-dimensional complex mode field must be jointly inferred from sparse observations. Their coupling makes the inference sensitive to mode-field representation errors and the optimization trajectory.

\subsection{Oscillatory Complex-Field Learning}
Standard multilayer perceptrons exhibit spectral bias, limiting their representation of localized and rapidly oscillatory solutions~\cite{rahaman2019spectral,xu2020frequency}. Fourier feature mappings improve access to high-frequency components by embedding spatial coordinates into a richer spectral space~\cite{tancik2020fourier}. Architectural~\cite{jagtap2020extended}, domain-decomposition, adaptive-sampling, loss-balancing, and optimization strategies further improve PINN expressivity and trainability~\cite{wu2023comprehensive,rathore2024challenges}. Complex-valued neural networks provide a foundation for learning in complex domains~\cite{trabelsi2018deep,zhang2025complex}. However, these advances remain largely separate in sparse complex-eigenfrequency inversion. CEI-PINN unifies complex-valued feature propagation, Fourier feature encoding, and staged optimization within a physics-informed framework.

\section{Preliminaries}

\subsection{Problem Formulation} \label{sec:problem_formulation}

We consider a large-aspect-ratio, circular-cross-section, axisymmetric
tokamak. Taking the rational surface \(r=r_0\) as reference, we define
\(x=r-r_0\), with \(\theta\) denoting the poloidal angle. The perturbed
electrostatic potential is
\begin{equation}
    \Phi(r,\theta,\zeta,t)
    =
    \phi(x,\theta)
    \mathrm{e}^{\mathrm{i}(n\zeta-\omega t)},
    \label{eq:perturbed_potential}
\end{equation}
where \(\zeta\) and \(n\) are the toroidal angle and mode number,
respectively. In the long-wavelength limit, \(\phi\) satisfies
\begin{equation}
    (L_0+L_1)\phi(x,\theta)=0,
    \label{eq:global_eigenvalue_equation}
\end{equation}
where
\begin{align}
    L_0={}&
    \rho_i^2\frac{\partial^2}{\partial x^2}
    -k_\theta^2\rho_i^2
    -
    \left(
        \frac{\omega_{*e}}{\omega}
        \frac{\epsilon_n}{qk_\theta\rho_i}
    \right)^2
    \left(
        \frac{\partial}{\partial\theta}
        +\mathrm{i}k_\theta sx
    \right)^2
    \nonumber\\
    &+
    \frac{
        \omega_{*e}-\omega(1-\mathrm{i}\delta_e)
    }{
        \omega_{*e}(1+\eta_i)/\tau+\omega
    },
    \label{eq:L0}\\
    L_1={}&
    -2\frac{\omega_{*e}}{\omega}\epsilon_n
    \left(
        \cos\theta
        +
        \sin\theta\frac{\mathrm{i}}{k_\theta}
        \frac{\partial}{\partial x}
    \right).
    \label{eq:L1}
\end{align}
Here, \(L_0\) describes radial, poloidal, magnetic-shear, diamagnetic,
and collisional effects, while \(L_1\) represents toroidal-curvature
coupling.

The equation admits multiple eigenmode branches, indexed radially and
poloidally by \(l_r\) and \(l_\theta\). We study the ground-state branch
\((l_r,l_\theta)=(0,0)\) on the computational domain \(\Omega\), whose
boundary is \(\partial\Omega\). Given \(N_d\) nonzero sparse observations
and \(N_b\) boundary samples,
\begin{equation}
    \mathcal{D}_d
    =
    \{(x_j^d,\theta_j^d,\phi_j^*)\}_{j=1}^{N_d},
    \qquad
    \mathcal{D}_b
    =
    \{(x_j^b,\theta_j^b,g_j)\}_{j=1}^{N_b},
\end{equation}
the inverse problem recovers a nontrivial field
\(\phi:\Omega\rightarrow\mathbb{C}\) and a global eigenfrequency
\(\omega\in\mathbb{C}\), The inverse problem seeks to recover a nontrivial complex mode field and a global complex eigenfrequency that satisfy the governing equation and
boundary conditions while agreeing with the sparse mode-field observations.

\begin{figure}[h]
    \centering
    \includegraphics[width=\linewidth]{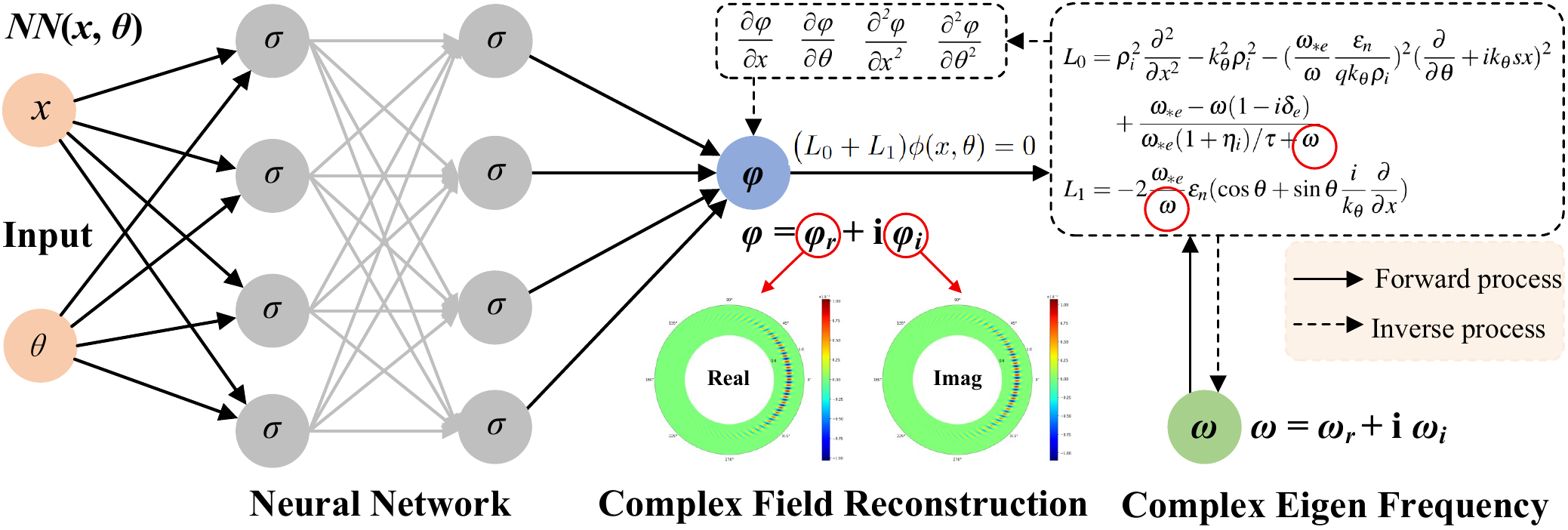}
    \caption{Schematic diagram of a PINN framework for solving a simplified 2D complex eigenvalue problem.}
    \label{fig:coupled_inversion}
\end{figure}

\subsection{Problem Interpretation.}
Equation~\eqref{eq:global_eigenvalue_equation} is a complex nonlinear
eigenvalue problem rather than an explicit input--output mapping. Its
unknowns are the mode field
\(\phi=\phi_r+\mathrm{i}\phi_i\) and eigenfrequency
\(\omega=\omega_r+\mathrm{i}\omega_i\), where \(\omega_r\) and
\(\omega_i\) are the oscillation frequency and growth rate. For each
physical case, the domain, boundary conditions, and parameters
\(\rho_i,k_\theta,q,s,\omega_{*e},\epsilon_n,\eta_i,\tau,\delta_e\)
are fixed. Since \(\omega\) enters \(L_0\) and \(L_1\) nonlinearly,
\(\phi\) and \(\omega\) must be recovered jointly. As illustrated in
Figure~\ref{fig:coupled_inversion}, a neural network approximates the
complex mode field from the spatial coordinates, while the governing
equation couples it to the global eigenfrequency. Moreover,
\(\phi\equiv0\) satisfies the homogeneous equation for any \(\omega\);
nonzero observations therefore select the target branch and exclude
this trivial solution.

\begin{figure*}[t]
    \centering
    \makebox[\textwidth][c]{%
        \includegraphics[
            width=\textwidth,
        ]{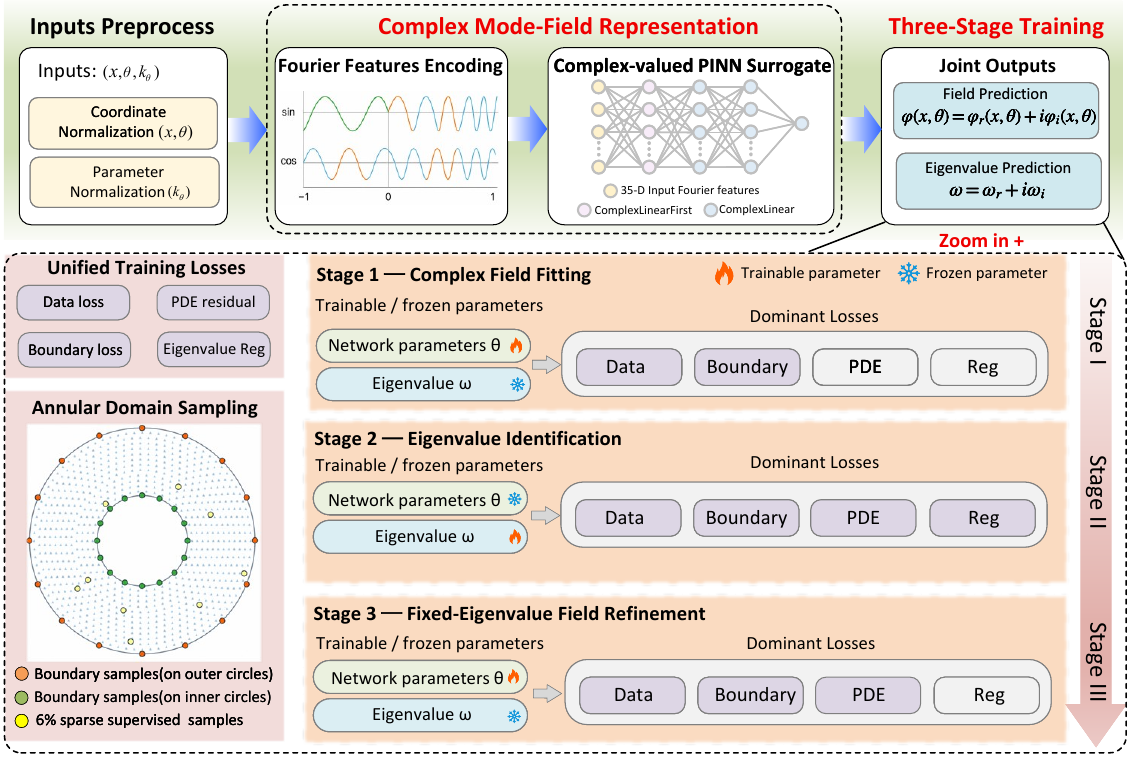}%
    }
    \caption{Overview of CEI-PINN for complex field reconstruction and eigenfrequency identification. Top: model architecture and three-stage training workflow; bottom: sparse observations and boundary samples in the annular domain.}
    \label{fig:model}
\end{figure*}

\section{The Proposed Method}

\subsection{Overall Architecture}
We propose CEI-PINN, a novel physics-constrained PINN for jointly
identifying the complex eigenfrequency and reconstructing the 2D complex
mode field. As shown in Figure~\ref{fig:model}, the normalized inputs
$(x,\theta,k_{\theta})$ are Fourier-encoded to capture oscillatory mode
structures~\cite{xu2020frequency,tancik2020fourier}. The field network
propagates these features while preserving real--imaginary interactions
and predicts $\phi_{\vartheta}$. Rather than treating the eigenfrequency
as a network output, CEI-PINN represents $\omega_r$ and $\omega_i$ as
independent trainable variables, with
$\omega=\omega_r+\mathrm{i}\omega_i$. The governing equation couples the
field and eigenfrequency and, together with sparse observations, boundary
conditions, and regularization, defines the data, boundary, PDE-residual,
and regularization losses. A three-stage schedule selectively freezes or
optimizes the field-network and eigenfrequency parameters, suppressing
compensatory solutions and stabilizing joint inversion.

\subsection{Complex Mode-Field Representation}

Accurate complex eigenfrequency identification requires both capturing the
rapid spatial variations of the mode field and modeling the coupling between
its real and imaginary components. CEI-PINN addresses these two requirements
through frequency-aware input encoding and complex-valued feature propagation,
respectively.
\subsubsection{Frequency-Aware Input Encoding}

The numerical ranges of the inputs \((x,\theta,k_{\theta})\) differ. Directly feeding the inputs into the network can cause gradient imbalance and alter the scale of the governing-equation residual. We first normalize the inputs as follows:
\textbf{1) Frequency-Aware Input Encoding.}
Since $(x,\theta,k_\theta)$ have different scales, we normalize them to
avoid gradient imbalance and residual rescaling:
\begin{equation}
\widehat{s}
=
2\frac{s-s_{\min}}{s_{\max}-s_{\min}}-1,
\quad s\in\{x,\theta\},
\qquad
\widehat{k}_{\theta}=\frac{k_{\theta}}{100}.
\label{eq:normalization}
\end{equation}
This transformation maps the spatial coordinates to $[-1,1]$ and scales the control parameter to a comparable range.

Although normalization improves numerical conditioning, it does not remove the
spectral bias of coordinate-based multilayer perceptrons. For each normalized
spatial coordinate $s\in\{\widehat{x},\widehat{\theta}\}$, we therefore use
the Fourier mapping
\begin{equation}
\begin{aligned}
&\gamma(s)=\Big[
s,\,
\sin(2^{0}\pi s),\cos(2^{0}\pi s),\,
\sin(2^{1}\pi s), \\
&\cos(2^{1}\pi s),\, \ldots,\,
\sin(2^{M-1}\pi s),\cos(2^{M-1}\pi s)
\Big].
\end{aligned}
\end{equation}

The mapping provides periodic bases at multiple spatial frequencies. The encoded coordinates are concatenated with the scaled control parameter to form the complex field-network input:
\begin{equation}
\mathbf{z}
=
\left[
\gamma(\widehat{x}),
\gamma(\widehat{\theta}),
\widehat{k}_{\theta}
\right].
\end{equation}

The resulting vector $\mathbf{z}$ serves as the input to the complex field network.

\subsubsection{Complex-Valued Feature Propagation}

Fourier encoding enhances spatial-frequency representation but does not explicitly couple the real and imaginary field components. We therefore propagate the encoded vector $\mathbf{z}$ through complex fully connected layers, which map and update the representation in a complex feature space. The final layer predicts the two components of the mode field:
\begin{equation}
\widehat{\phi}_{\vartheta}(x,\theta,k_{\theta})
=
\widehat{\phi}_{r}(x,\theta,k_{\theta})
+
\mathrm{i}\widehat{\phi}_{i}(x,\theta,k_{\theta}).
\label{eq:complex_field_prediction}
\end{equation}

This design enables real--imaginary interaction throughout feature propagation rather than only through the final loss. 

\subsection{Physics-Informed Learning Objective}
\label{sec:learning_objective}

The learning objective combines sparse observations, boundary conditions, the global ITG eigenvalue equation, and eigenfrequency regularization.

\paragraph{Data supervision}
Given $N_d$ sparse observations
$\{(x_j^d,\theta_j^d,\phi_j^{*})\}_{j=1}^{N_d}$, the data loss is
\begin{equation}
    \mathcal{L}_{\mathrm{data}}
    =
    \frac{1}{N_d}
    \sum_{j=1}^{N_d}
    \left\|
        \phi_{\vartheta}
        (x_j^d,\theta_j^d)
        -
        \phi_j^{*}
    \right\|_2^2.
    \label{eq:data_loss_method}
\end{equation}
This term aligns the predicted field with the observed mode.

\paragraph{Boundary Constraint}
The annular domain has inner and outer boundaries, with
$\partial\Omega=\partial\Omega_{\mathrm{in}}\cup\partial\Omega_{\mathrm{out}}$.
We impose the Dirichlet condition
\begin{equation}
    \phi(x,\theta)
    =
    g(x,\theta),
    \qquad
    (x,\theta)\in\partial\Omega,
    \label{eq:dirichlet_condition}
\end{equation}
where $g(x,\theta)=0$ in our setting. For $N_b$ boundary samples, the
corresponding loss is
\begin{equation}
    \mathcal{L}_{\mathrm{bc}}
    =
    \frac{1}{N_b}
    \sum_{j=1}^{N_b}
    \left\|
        \phi_{\vartheta}
        (x_j^b,\theta_j^b)
    \right\|_2^2.
    \label{eq:boundary_loss_method}
\end{equation}

All remaining coefficients in $L_0$ and $L_1$ are fixed physical parameters
of the global fluid model. Substituting the predicted field gives
\begin{equation}
    \mathcal{R}
    \left(
        x,\theta;
        \phi_{\vartheta},
        \omega
    \right)
    =
    \big(L_0+L_1)
    \phi_{\vartheta}
    (x,\theta).
    \label{eq:itg_residual}
\end{equation}
For $N_f$ interior collocation points, the PDE loss is
\begin{equation}
    \mathcal{L}_{\mathrm{pde}}
    =
    \frac{1}{N_f}
    \sum_{j=1}^{N_f}
    \left\|
        \mathcal{R}
        \left(
            x_j^f,\theta_j^f;
            \phi_{\vartheta},
            \omega
        \right)
    \right\|_2^2.
    \label{eq:pde_loss_method}
\end{equation}

\paragraph{Frequency Regularization}
We use a soft regularizer to restrict the eigenfrequency to a physically admissible range:
\begin{equation}
    \mathcal{L}_{\mathrm{reg}}
    =
    \mathcal{R}_{\omega}(\omega).
    \label{eq:eigenfrequency_regularization}
\end{equation}
The complete objective is
\begin{equation}
    \mathcal{L}
    =
    \mathcal{L}_{\mathrm{data}}
    +
    \mathcal{L}_{\mathrm{bc}}
    +
    \mathcal{L}_{\mathrm{pde}}
    +
    \mathcal{L}_{\mathrm{reg}},
    \label{eq:complete_objective}
\end{equation}
where the four coefficients balance the constraints.

\subsection{Three-Stage Training}

Jointly optimizing $\vartheta$ and $\omega$ in Eq.~\eqref{eq:complete_objective} allows the field network to compensate for eigenfrequency errors by altering the mode structure. CEI-PINN mitigates this coupling through three successive training stages.

\paragraph{Stage I: Mode-field fitting}
We fix $\omega$ at its initial value and optimize only $\vartheta$ for
$15{,}000$ epochs. For $0\leq e<2{,}000$, the network is trained using the
data and boundary losses. The PDE loss is activated at epoch $2{,}000$ and
retained for the remainder of Stage~I:
\begin{equation}
\begin{aligned}
\mathcal{L}_{\mathrm{I}}(e)
={}&
\mathcal{L}_{\mathrm{data}}
+
\mathcal{L}_{\mathrm{bc}}
+
\begin{cases}
0,
& 0 \leq e < 2{,}000,
\\[3pt]
\mathcal{L}_{\mathrm{pde}},
& 2{,}000 \leq e < 15{,}000.
\end{cases}
\end{aligned}
\label{eq:stage_I}
\end{equation}
where \(e\) is the training epoch. The warm-up anchors mode-field amplitude and phase using sparse data and boundary constraints; then PDE enforcement improves its physical consistency before eigenfrequency identification.

\paragraph{Stage II: Complex eigenfrequency identification}
We freeze the field network and optimize only $\omega_r$ and $\omega_i$:
\begin{equation}
    \mathcal{L}_{\mathrm{II}}
    =
    \mathcal{L}_{\mathrm{data}}
    +
    \mathcal{L}_{\mathrm{pde}}
     +
    \mathcal{L}_{\mathrm{bc}}
    +
    \mathcal{L}_{\mathrm{reg}}.
    \label{eq:stage_two_objective}
\end{equation}
More explicitly, with $\vartheta=\vartheta^{(1)}$ fixed, the eigenfrequency identification problem and the identified complex eigenfrequency are
\begin{equation}
(\widehat{\omega}_r,\widehat{\omega}_i)
=
\operatorname*{arg\,min}_{\omega_r,\omega_i}
\mathcal{L}_{\mathrm{II}},
\qquad
\widehat{\omega}
=
\widehat{\omega}_r+\mathrm{i}\widehat{\omega}_i.
\label{eq:stage_II_identification}
\end{equation}

Fixing the observation-consistent field prevents changes in the mode structure from offsetting eigenfrequency errors.

\paragraph{Stage III: Fixed-eigenfrequency field refinement}
After the complex eigenfrequency converges, we fix it and fine-tune the field
network with
\begin{equation}
    \mathcal{L}_{\mathrm{III}}
    =
    \mathcal{L}_{\mathrm{data}}
    +
    \mathcal{L}_{\mathrm{bc}}
    +
    \mathcal{L}_{\mathrm{pde}}.
    \label{eq:stage_three_objective}
\end{equation}
This stage improves agreement between the reconstructed field and the
governing equation without altering $\widehat{\omega}$. Overall, the staged
procedure reduces gradient interference and progressively separates field
initialization, eigenfrequency identification, and physics-informed field
refinement.
\section{Experiments} 

\subsection{Datasets and Evaluation Metrics} 

\paragraph{Datasets}
Plasma physicists generated reference data for the steep-gradient ITG ground-state branch $(l_r,l_\theta)=(0,0)$ using a custom numerical eigensolver, yielding the ground-truth complex eigenfrequency and mode field. Training on the annular domain uses sparse field observations, boundary samples enforcing the boundary conditions, and PDE collocation points evaluating the PDE residual. These data and constraints enable joint eigenfrequency identification and complex-valued mode-field reconstruction.

\paragraph{Evaluation Metrics}
Eigenfrequency recovery is quantified by the component-wise absolute errors
$E_r=|\widehat{\omega}_r-\omega_r^{\mathrm{ref}}|$ and
$E_i=|\widehat{\omega}_i-\omega_i^{\mathrm{ref}}|$, together with the overall
relative error
$E_{\omega}
=
|\widehat{\omega}-\omega^{\mathrm{ref}}|
/
|\omega^{\mathrm{ref}}|$,
where
$\widehat{\omega}
=
\widehat{\omega}_r+\mathrm{i}\widehat{\omega}_i$
and
$\omega^{\mathrm{ref}}
=
\omega_r^{\mathrm{ref}}+\mathrm{i}\omega_i^{\mathrm{ref}}$.
Here, $E_r$ and $E_i$ measure the recovery errors of the oscillation frequency
and linear growth rate, respectively. Mode-field reconstruction is evaluated using the aligned relative $L_2$ error,
$E_{\varphi}^{\mathrm{align}}
=
\min_{c\in\mathbb{C}}
\|c\widehat{\boldsymbol{\varphi}}-\boldsymbol{\varphi}^{*}\|_2
/
\|\boldsymbol{\varphi}^{*}\|_2$.
Here, $\widehat{\boldsymbol{\varphi}}$ and
$\boldsymbol{\varphi}^{*}$ are the predicted and reference complex mode fields over the evaluation grid, and $c$ aligns their global amplitude and phase. Together, these metrics assess eigenfrequency recovery and mode-field reconstruction.

\subsection{Implementation Details}
CEI-PINN maps the normalized, Fourier-encoded spatial coordinates
\((x,\theta)\) and the control parameter \(k_\theta\) to the complex mode
field \(\widehat{\phi}\) through a complex-valued neural network. The Fourier
feature encoder uses eight frequency bands, producing a 35-dimensional input
after concatenation with the normalized \(k_\theta\). The field network
contains three hidden complex-valued layers, each with 128 neurons and
\(\tanh\) activation. The real and imaginary components, \(\omega_r\) and
\(\omega_i\), of the complex eigenfrequency
\(\omega=\omega_r+\mathrm{i}\omega_i\) are represented as independent
trainable parameters.

Training uses \(6\%\) randomly sampled mode-field observations, \(80{,}000\) 
PDE collocation points, and \(16{,}000\) boundary points. The model is optimized 
for \(40{,}000\) epochs using Adam and a three-stage training strategy. In
Stage I, the complex eigenfrequency is fixed and the field network is trained
with a learning rate of \(5\times10^{-3}\). In Stage II, the field network is
frozen and the two eigenfrequency components are optimized with a learning
rate of \(1\times10^{-3}\). In Stage III, the identified eigenfrequency is
fixed and the field network is refined with a learning rate of
\(1\times10^{-5}\). Gradient clipping with a maximum norm of \(5.0\) is
applied during training, and the random seed is set to \(42\). All experiments
are conducted on a single NVIDIA GeForce RTX 4090 GPU.


\subsection{Main Results} 
\paragraph{Comparison with PINN Backbones}
We compare CEI-PINN with four representative architectures: vanilla PINN, PINNFormer, PINNMamba, and PINKAN. All methods are evaluated for complex eigenfrequency identification under identical sparse supervision. As shown in Table~\ref{tab:pinn_backbone_comparison}, CEI-PINN converges to \(\widehat{\omega}=2.175+2.927\mathrm{i}\) and achieves the lowest
errors across all frequency metrics: \(E_r=0.087\), \(E_i=0.059\), and \(E_\omega=0.029\).
In particular, CEI-PINN reduces the overall relative error by \(79.6\%\) compared with the best-performing baseline, vanilla PINN (\(E_\omega=0.142\)). These results indicate that CEI-PINN more accurately recovers both eigenfrequency components under limited observations.

\begin{table}[h]
\centering
\caption{Comparison of PINN architectures for complex eigenfrequency identification.}
\label{tab:pinn_backbone_comparison}
\resizebox{\columnwidth}{!}{
\begin{tabular}{lcccc}
\toprule
\toprule
\textbf{Method} & \textbf{Converged $\omega$}
& $\bm{E_r}$ & $\bm{E_i}$ & $\bm{E_{\omega}}$ \\
\midrule
PINN       & $2.000 + 3.496i$ & 0.088 & 0.510 & 0.142 \\
PINNFormer & $3.545 + 3.055i$ & 1.459 & 0.069 & 0.401 \\
PINNMamba  & $7.772 + 3.463i$ & 5.684 & 0.476 & 1.565 \\
PINKAN     & $0.849 + 0.851i$ & 1.239 & 2.136 & 0.678 \\
Ours       & $\mathbf{2.175 + 2.927i}$ & $\mathbf{0.087}$
& $\mathbf{0.059}$ & $\mathbf{0.029}$ \\
\bottomrule
\end{tabular}
}
\end{table}

\paragraph{Performance under Different Supervision Ratios}
Table~\ref{tab:supervision_ratio} reports the identification and reconstruction results for supervision ratios of $1\%$--$10\%$. The framework recovers near-reference eigenfrequencies in several settings, achieving its best overall performance at $6\%$. It converges to $\widehat{\omega}=2.175+2.927\mathrm{i}$, with relative eigenfrequency and aligned relative $L_2$ errors of $0.03$ and $0.18$, respectively. This demonstrates joint eigenfrequency identification and complex-valued mode-field reconstruction from sparse observations. Performance is non-monotonic: the reconstruction error decreases from $0.59$ to $0.18$ between $1\%$ and $6\%$, whereas the eigenfrequency error rises to approximately $0.13$--$0.14$ at $7\%$, $8\%$, and $10\%$. This trend reflects the interplay among data quantity, parameter coupling, optimization trajectory, and physical constraints, indicating that balancing observations and physics is more important than simply increasing supervision.


\begin{table}[h]
\centering
\caption{Performance under different supervision ratios for complex eigenfrequency
identification and complex mode-field reconstruction.}
\label{tab:supervision_ratio}

\small
\setlength{\tabcolsep}{3pt}
\resizebox{\columnwidth}{!}{%
\begin{tabular}{c c c c c c}
\toprule
\textbf{Ratio}
& \textbf{Converged $\omega$}
& $\bm{E_r}$
& $\bm{E_i}$
& $\bm{E_{\omega}}$
& $\bm{E_{\varphi}^{\mathrm{align}}}$ \\
\midrule
1\%  & $2.169 + 3.498\mathrm{i}$ & $0.081$ & $0.512$ & $0.14$ & $0.59$ \\
2\%  & $2.168 + 3.486\mathrm{i}$ & $0.080$ & $0.500$ & $0.14$ & $0.39$ \\
3\%  & $2.168 + 2.927\mathrm{i}$ & $0.080$ & $0.059$ & $0.03$ & $0.35$ \\
4\%  & $2.172 + 2.931\mathrm{i}$ & $0.084$ & $0.055$ & $0.03$ & $0.31$ \\
5\%  & $2.175 + 2.930\mathrm{i}$ & $0.087$ & $0.056$ & $0.03$ & $0.30$ \\
\textbf{6\%}
& $\mathbf{2.175 + 2.927\mathrm{i}}$
& $0.087$
& $0.059$
& $\mathbf{0.03}$
& $\mathbf{0.18}$ \\
7\%  & $2.183 + 3.455\mathrm{i}$ & $0.095$ & $0.469$ & $0.13$ & $0.25$ \\
8\%  & $2.178 + 3.492\mathrm{i}$ & $0.090$ & $0.506$ & $0.14$ & $0.27$ \\
9\%  & $2.172 + 2.931\mathrm{i}$ & $0.084$ & $0.055$ & $0.03$ & $0.28$ \\
10\% & $2.194 + 3.469\mathrm{i}$ & $0.106$ & $0.483$ & $0.14$ & $0.24$ \\
\bottomrule
\end{tabular}
}
\end{table}

\subsection{Ablation Studies}

\paragraph{Effect of Fourier Feature Encoding}
Table~\ref{tab:fourier_ablation} summarizes the effect of the number of Fourier frequencies $n_f$. Without Fourier encoding, the relative eigenfrequency error is $0.0368$, while $n_f=2$ and $n_f=4$ yield imaginary-part errors of $0.5036$ and $0.4874$, respectively. The lowest relative eigenfrequency error, $0.0289$, occurs at $n_f=8$, but increases to $0.0389$ at $n_f=16$. Thus, $n_f=8$ is selected as the default, balancing oscillatory-field representation and coupled-optimization difficulty.

\begin{table}[h]
\centering
\caption{Fourier-frequency ablation for complex eigenfrequency identification.}
\label{tab:fourier_ablation}
\resizebox{\columnwidth}{!}{%
\begin{tabular}{lcccc}
\toprule
\toprule
\textbf{Ablation}
& \textbf{Converged }$\bm{\omega}$
& $\bm{E_r}$
& $\bm{E_i}$
& $\bm{E_{\omega}}$ \\
\midrule
No Fourier
& $2.193+2.904\mathrm{i}$
& 0.106 & 0.082 & 0.037 \\

Fourier ($n_f=2$)
& $2.168+3.490\mathrm{i}$
& 0.081 & 0.504 & 0.140 \\

Fourier ($n_f=4$)
& $2.170+3.474\mathrm{i}$
& 0.083 & 0.487 & 0.136 \\

Fourier ($n_f=8$)
& $\mathbf{2.175+2.927\mathrm{i}}$
& 0.087
& $\mathbf{0.059}$
& $\mathbf{0.029}$ \\

Fourier ($n_f=16$)
& $2.200+2.900\mathrm{i}$
& 0.112 & 0.086 & 0.039 \\
\bottomrule
\end{tabular}%
}
\end{table}

\paragraph{Effect of Key Representation Components}
Table~\ref{tab:ablation-components} reports the progressive ablation of Fourier feature encoding and complex-valued feature propagation. The full model converges to $\widehat{\omega}=2.175+2.927\mathrm{i}$, achieving the lowest eigenfrequency error of $0.029$. Removing complex-valued feature propagation increases the error to $0.128$, and further removing Fourier feature encoding increases it to $0.142$. These results demonstrate that both components contribute to complex eigenfrequency identification and
that their combination yields the best performance.









\paragraph{Effect of the Three-Stage Training Strategy}
Table~\ref{tab:training_strategy_ablation} compares joint, alternating, and proposed three-stage training. Joint training converges to $\widehat{\omega}=2.137+3.494\mathrm{i}$ with $E_{\omega}=0.139$, whereas alternating training reduces the error to $0.038$. The three-stage strategy performs best, converging to $\widehat{\omega}=2.175+2.927\mathrm{i}$ with $E_{\omega}=0.029$. Separating mode-field initialization, eigenfrequency identification, and fixed-eigenfrequency field refinement reduces coupled-optimization difficulty and improves accuracy.

\begin{table}[h]
\centering
\caption{Ablation of representation components for complex eigenfrequency identification. Checkmarks denote inclusion.}
\label{tab:ablation-components}

\setlength{\tabcolsep}{4pt}
\renewcommand{\arraystretch}{1.12}

\resizebox{\columnwidth}{!}{%
\begin{tabular}{cccccc}
\toprule
\toprule
\textbf{Fourier}
& \textbf{Complex}
& \textbf{Converged }$\bm{\omega}$
& $\bm{E_r}$
& $\bm{E_i}$
& $\bm{E_{\omega}}$ \\
\midrule

$\checkmark$
& $\checkmark$
& $\bm{2.175+2.927\mathrm{i}}$
& $\bm{0.087}$
& $\bm{0.059}$
& $\bm{0.029}$ \\

$\checkmark$
& --
& $2.193+3.441\mathrm{i}$
& $0.105$
& $0.455$
& $0.128$ \\

--
&$\checkmark$
& $2.193+2.904\mathrm{i}$
& $0.106$
& $0.082$
& $0.037$ \\

--
& --
& $2.000+3.496\mathrm{i}$
& $0.088$
& $0.510$
& $0.142$ \\

\bottomrule
\end{tabular}%
}
\end{table}
\begin{table}[h]
\centering
\caption{Training-strategy ablation for complex eigenfrequency identification.}
\label{tab:training_strategy_ablation}

\resizebox{\columnwidth}{!}{%
\begin{tabular}{lcccc}
\toprule
\toprule
\textbf{Method}
& \textbf{Converged }$\bm{\omega}$
& $\bm{E_r}$
& $\bm{E_i}$
& $\bm{E_{\omega}}$ \\
\midrule

Joint
& $2.137+3.494\mathrm{i}$
& $\bm{0.049}$
& $0.507$
& $0.139$ \\

Alternating
& $2.200+2.900\mathrm{i}$
& $0.112$
& $0.087$
& $0.038$ \\

Three-stage (Ours)
& $\bm{2.175+2.927\mathrm{i}}$
& $0.087$
& $\bm{0.059}$
& $\bm{0.029}$ \\

\bottomrule
\end{tabular}%
}
\end{table}

\paragraph{Effect of Loss Components.}
Additional ablations show that the PDE residual couples the mode field and complex eigenfrequency, while data, boundary, and regularization terms support target-mode selection and inversion stability. 

\begin{table}[h]
    \centering
    \caption{Convergence Analysis of $\omega$ Initial Values}
    \label{tab:omega_convergence}
    \resizebox{\columnwidth}{!}{%
    \begin{tabular}{cc|cc} 
        \toprule
        \toprule
        \textbf{Initial} \textbf{$\omega$} & \textbf{Converged Value} & \textbf{Initial $\omega$} & \textbf{Converged Value} \\
        \midrule
        $1+1i$   & $2.185+2.904i$ & $3+2.5i$ & $2.174+2.919i$ \\
        $1+1.5i$ & $2.168+2.908i$ & $3+3i$   & $2.197+2.931i$ \\
        $1+2i$   & $2.184+2.931i$ & $3+3.5i$ & $2.199+2.931i$ \\
        $1+2.5i$ & $2.174+2.904i$ & $4+1i$   & $2.186+2.931i$ \\
        $2+1i$   & $2.169+2.931i$ & $4+1.5i$ & $2.180+2.929i$ \\
        $2+1.5i$ & $2.175+2.911i$ & $4+2i$   & $2.178+2.930i$ \\
        $2+2i$   & $2.169+2.907i$ & $4+2.5i$ & $2.173+2.930i$ \\
        $2+2.5i$ & $2.177+2.922i$ & $4+3i$   & $2.168+2.931i$ \\
        $2+3i$   & $2.179+2.930i$ & $5+1i$   & $2.191+2.908i$ \\
        $2+3.5i$ & $2.175+2.927i$ & $5+1.5i$ & $2.196+2.905i$ \\
        $3+1i$   & $2.173+2.901i$ & $5+2i$   & $2.192+2.906i$ \\
        $3+1.5i$ & $2.179+2.905i$ & $5+2.5i$ & $2.199+2.931i$ \\
        $3+2i$   & $2.171+2.927i$ & $5+3i$   & $2.182+2.920i$ \\
        \bottomrule
    \end{tabular}%
    }
\end{table}

\paragraph{Robustness to the Initial Complex Eigenfrequency.}
Table~\ref{tab:omega_convergence} evaluates CEI-PINN sensitivity to different $\omega$ initializations within the specified search range. The model consistently converges near the target eigenfrequency, indicating reduced initialization sensitivity and improved stability. The remaining variations reflect interactions among the initialization of the eigenfrequency, the learning of the mode-field, and the residual constraints of the PDE.

\subsection{Visualization}
\begin{figure}[!t]
\centering
\setlength{\tabcolsep}{1pt}

\newcommand{\rowlabel}[1]{%
  \makebox[0.05\linewidth][c]{%
    \rotatebox{90}{\scriptsize #1}%
  }%
}

\newcommand{\colhead}[1]{%
  \hspace{-9pt}
  \makebox[0.29\linewidth][c]{\scriptsize #1}%
}

\newcommand{\panel}[1]{%
  \IfFileExists{#1}{%
    \includegraphics[width=0.29\linewidth]{#1}%
  }{%
    \fbox{\parbox[c][0.18\linewidth][c]{0.27\linewidth}{%
      \centering\scriptsize Missing image%
    }}%
  }%
}

\begin{tabular}{@{}c c c c@{}}
&
\colhead{Prediction} &
\colhead{Ground Truth} &
\colhead{Error}
\\[-3pt]

\raisebox{16pt}{\rowlabel{Real Part}} &
\panel{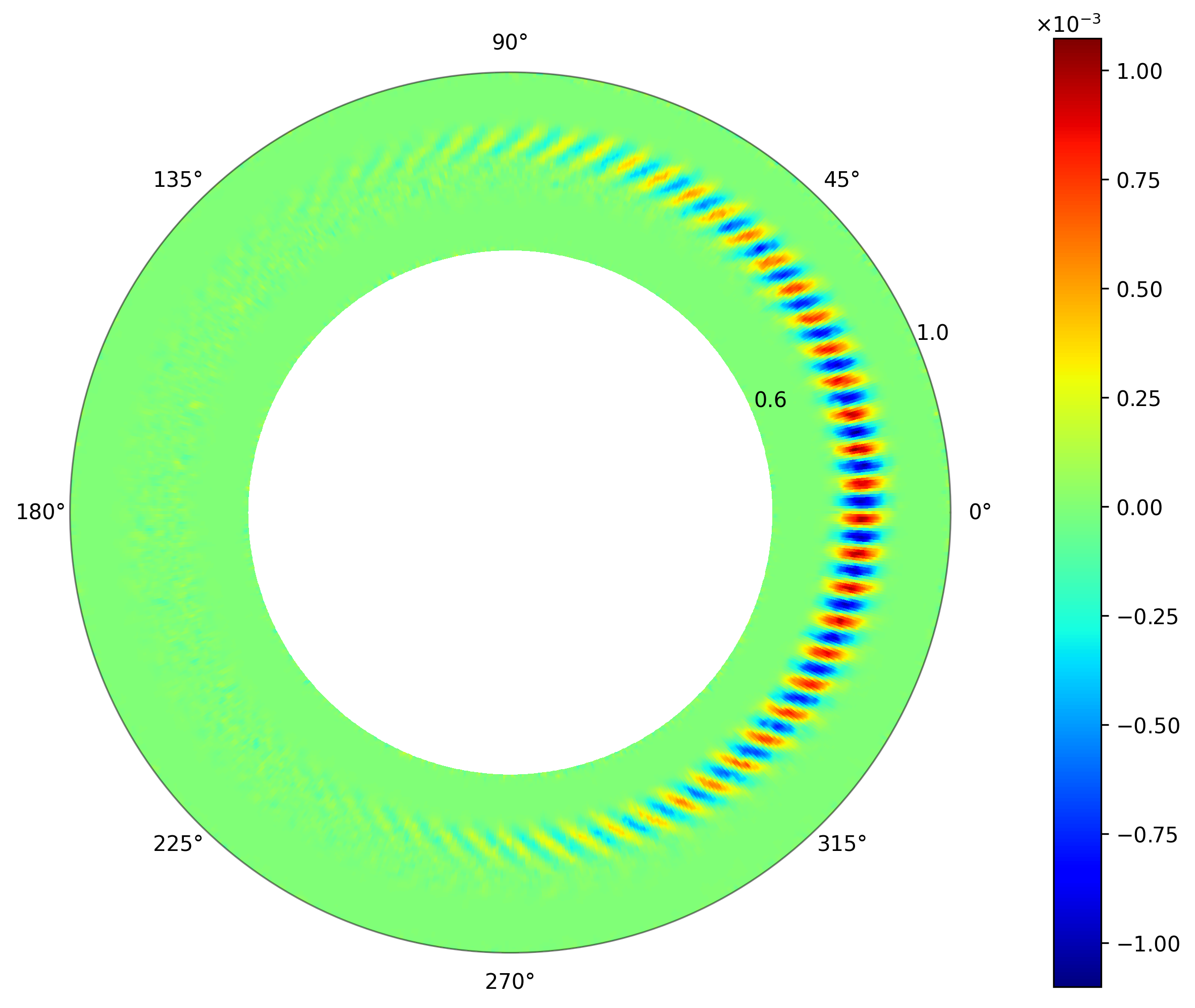} &
\panel{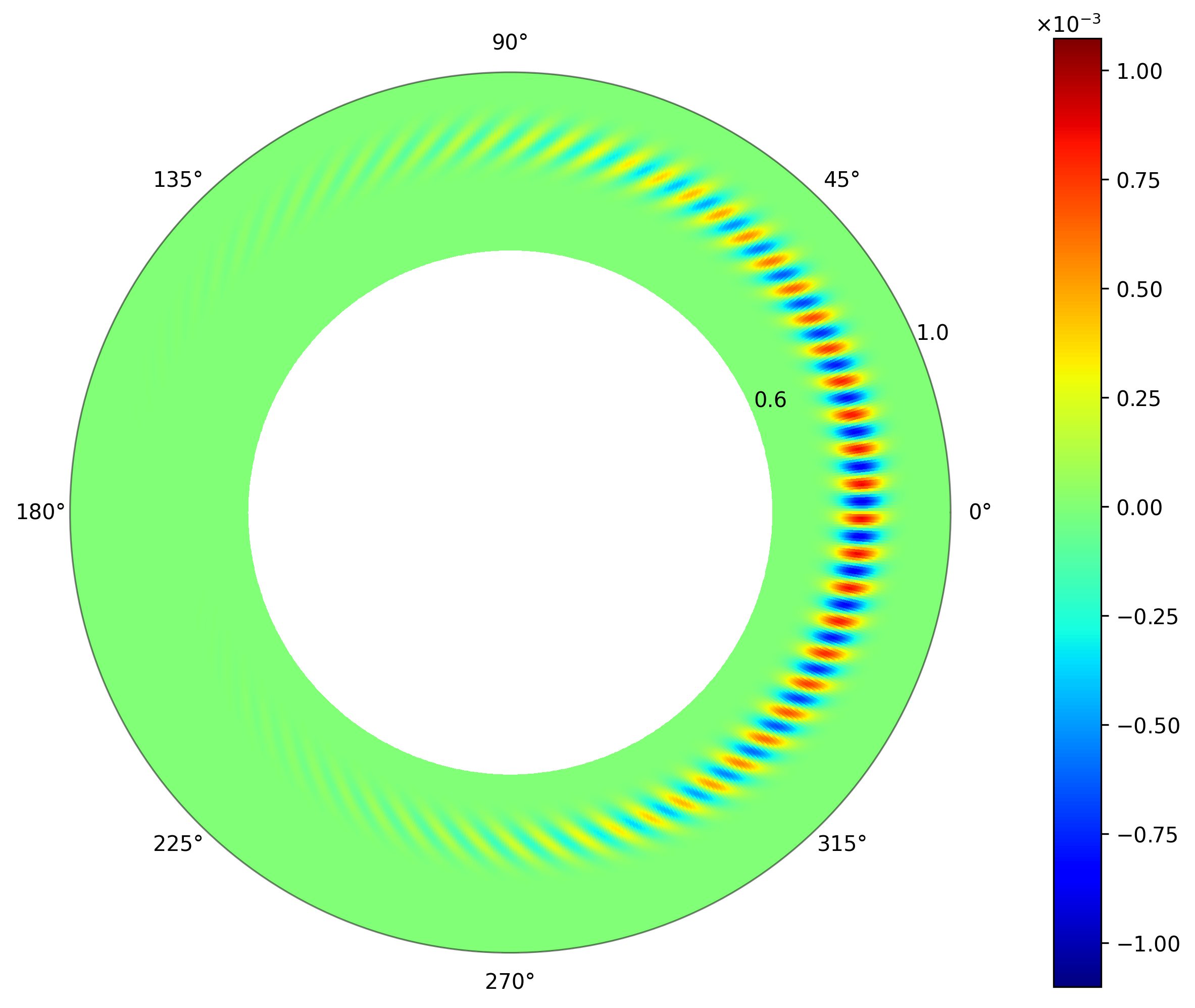} &
\panel{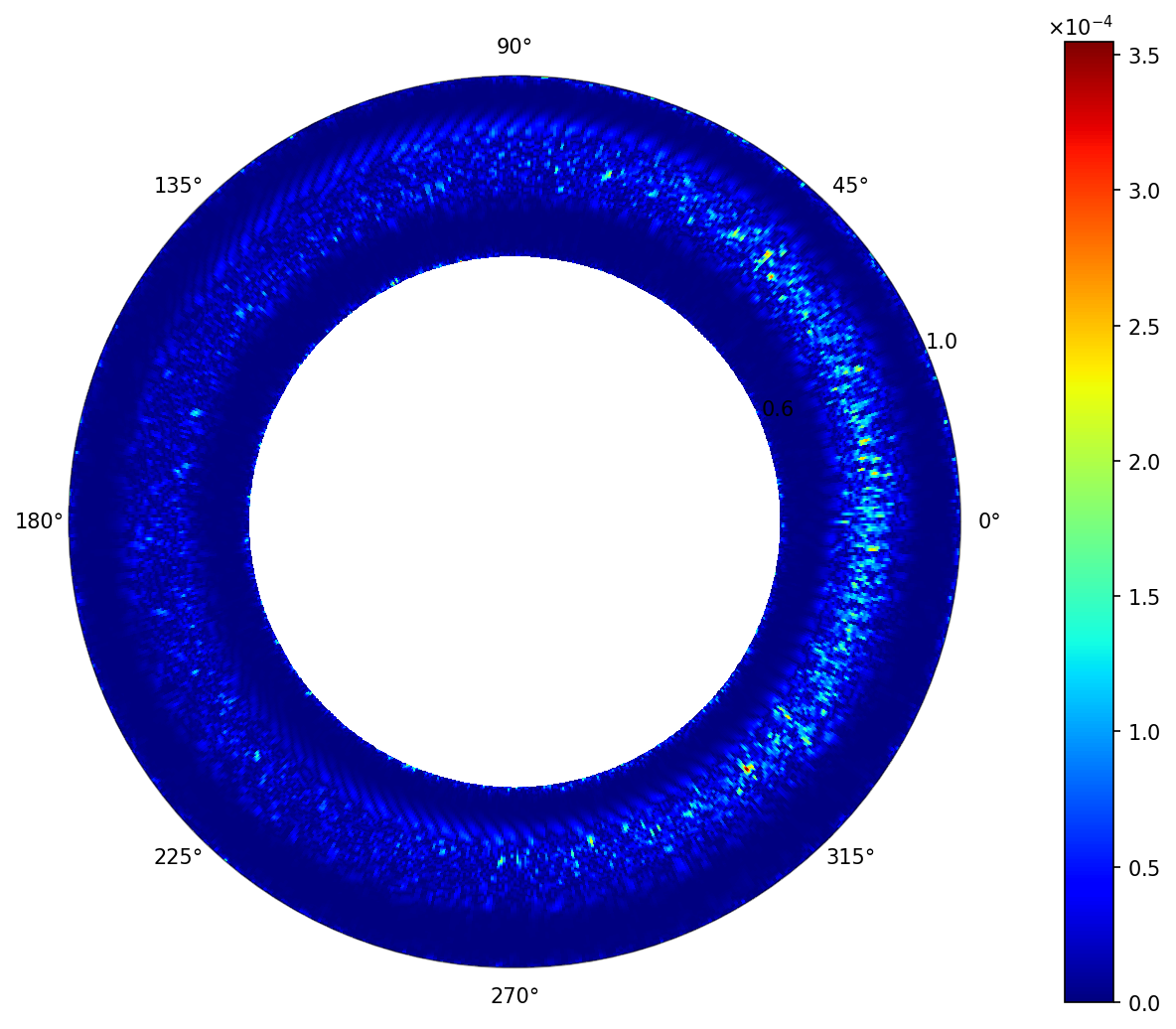}
\\[3pt]

\raisebox{16pt}{\rowlabel{Imag Part}} &
\panel{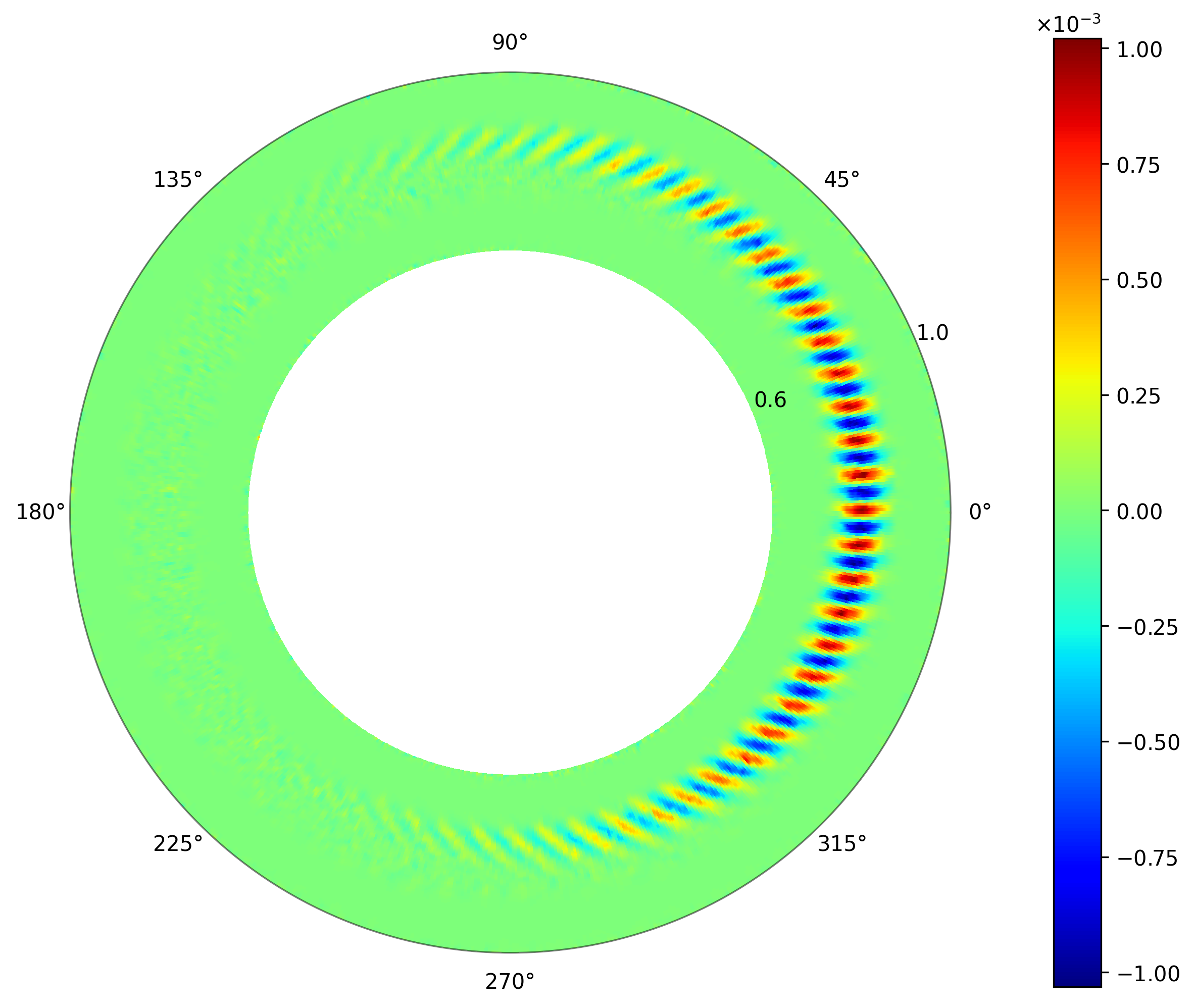} &
\panel{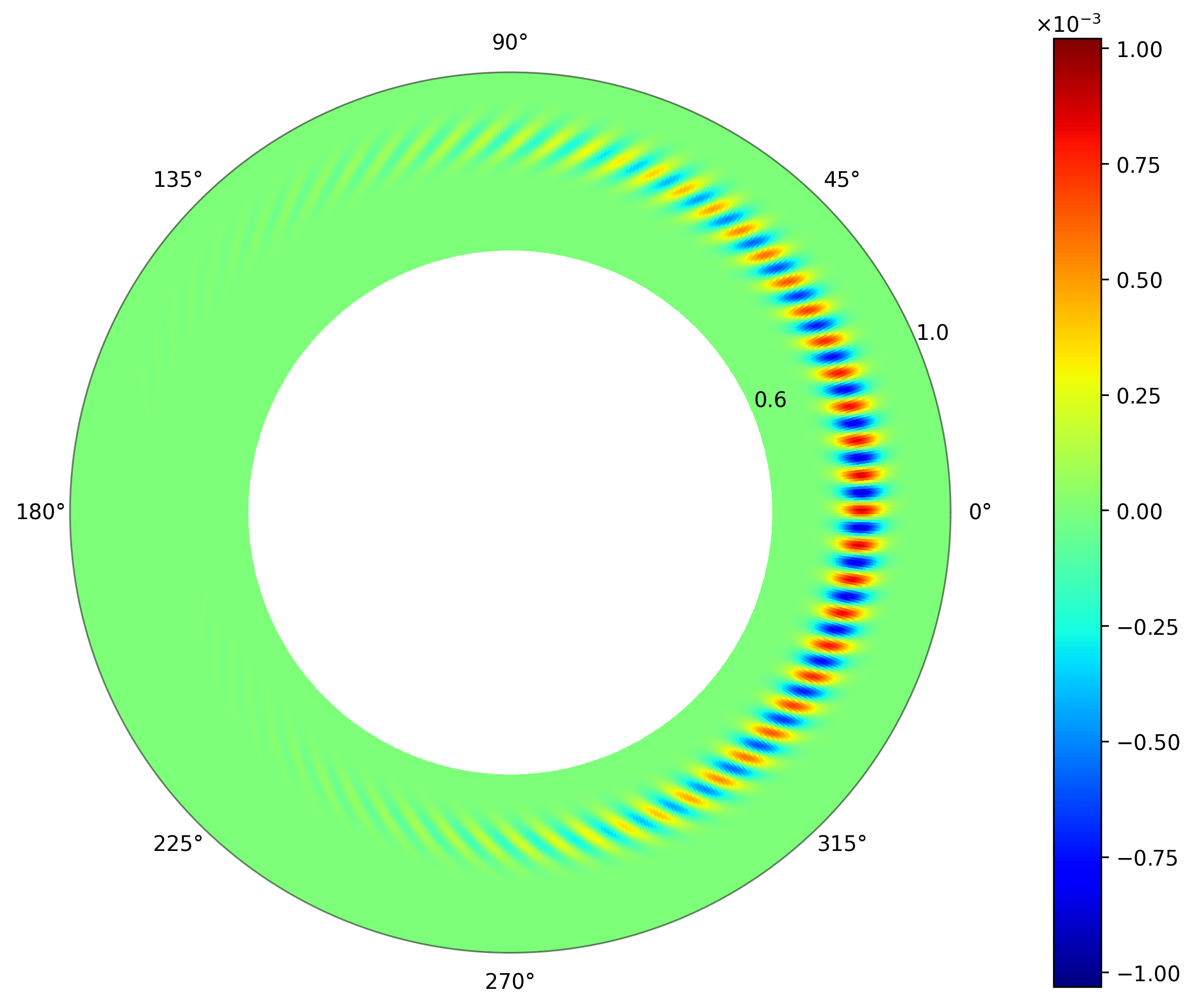} &
\panel{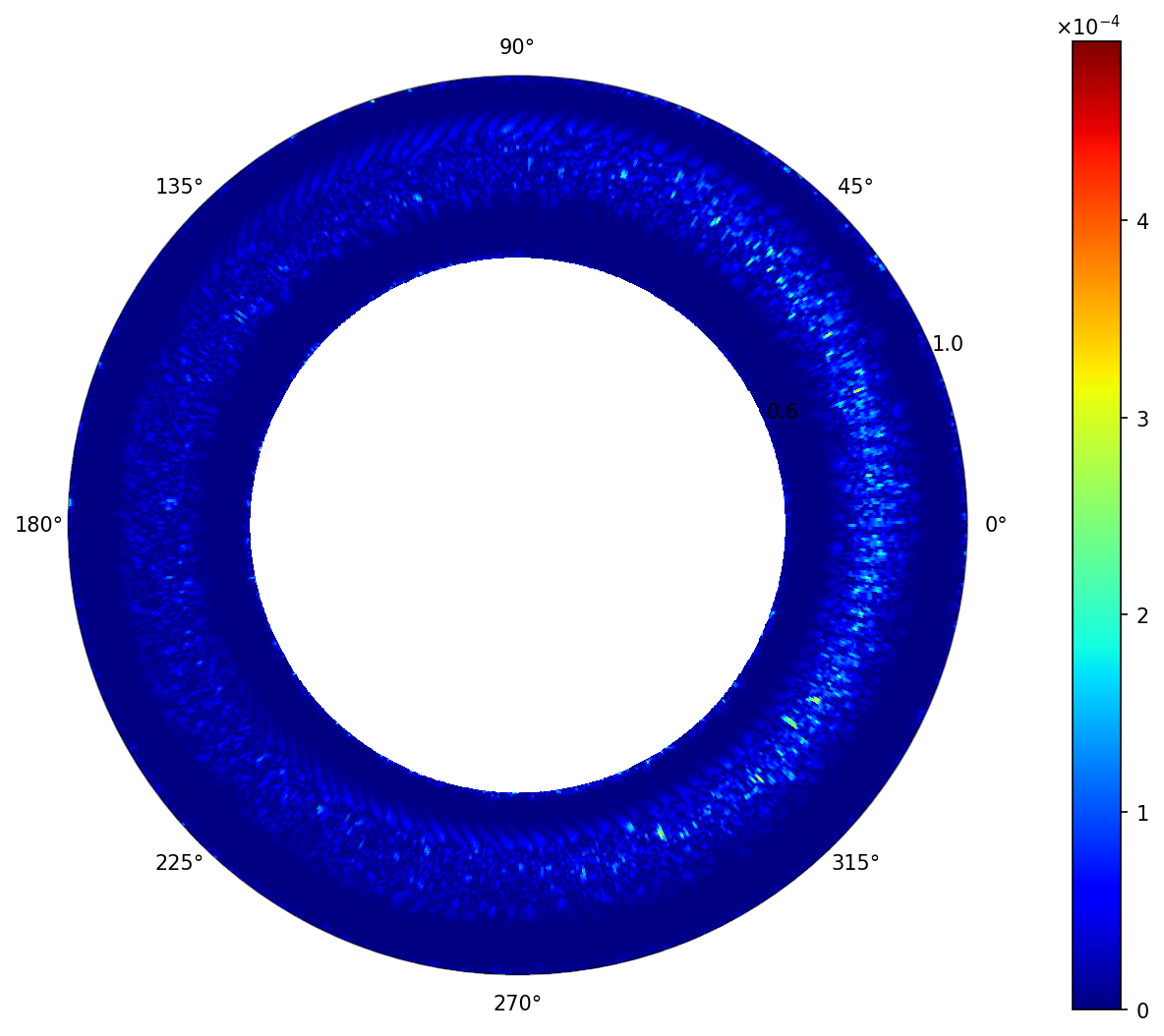}
\end{tabular}

\caption{Complex-valued mode-field reconstruction under 6\% supervision.}
\label{fig:complex_field_vis}
\end{figure}

Figure~\ref{fig:complex_field_vis} shows the reconstructed complex mode field under $6\%$ supervision. The predicted real and imaginary components reproduce the dominant spatial distribution, poloidal oscillations, peak locations, and overall contours of the reference field. Errors are concentrated in regions with rapid structural variation and remain low over most of the domain, consistent with the quantitative result in Table~\ref{tab:supervision_ratio}. Together, the visual and quantitative results confirm accurate complex mode field reconstruction.

\subsection{Limitation Analysis}

Experiments focus on complex eigenfrequency identification and complex mode-field reconstruction for the representative ITG ground-state branch,
showing that the framework works well in this setting. Future work may extend it to higher-order and multiple branches with more complex modes. The method uses sparse supervision and physics constraints to guide mode recovery. Efficient sampling and adaptive constraints may reduce reliance on manual
parameter settings across physical conditions. Residual-driven sampling and branch-aware optimization may further improve generalization while preserving the stability provided by eigenfrequency regularization.

\section{Conclusion} \label{sec:conclusion}

This work introduces CEI-PINN to jointly identify complex eigenfrequencies and reconstruct 2D mode fields from sparse observations of steep-gradient ITG modes. It combines Fourier feature encoding, complex-valued feature propagation, and three-stage optimization to reduce their coupling. For the ground-state branch, CEI-PINN achieves \(E_\omega=0.029\) and \(E_\phi^{\mathrm{align}}=0.18\) under \(6\%\) supervision, outperforming four PINN baselines in eigenfrequency error. Ablation and initialization studies support the proposed representation and training strategy. However, performance varies non-monotonically with the supervision ratio, showing that more sparse observations do not necessarily improve inversion accuracy. Validation is limited to the ground-state branch under prescribed physical conditions; extensions to multiple branches and adaptive constraints remain future work.

\bibliographystyle{IEEEtran}
\bibliography{references}

\appendices 
\clearpage 

\section{Physics of the ITG Inverse Problem}
\begin{figure}[htbp]
    \centering
    \includegraphics[width=0.75\linewidth]{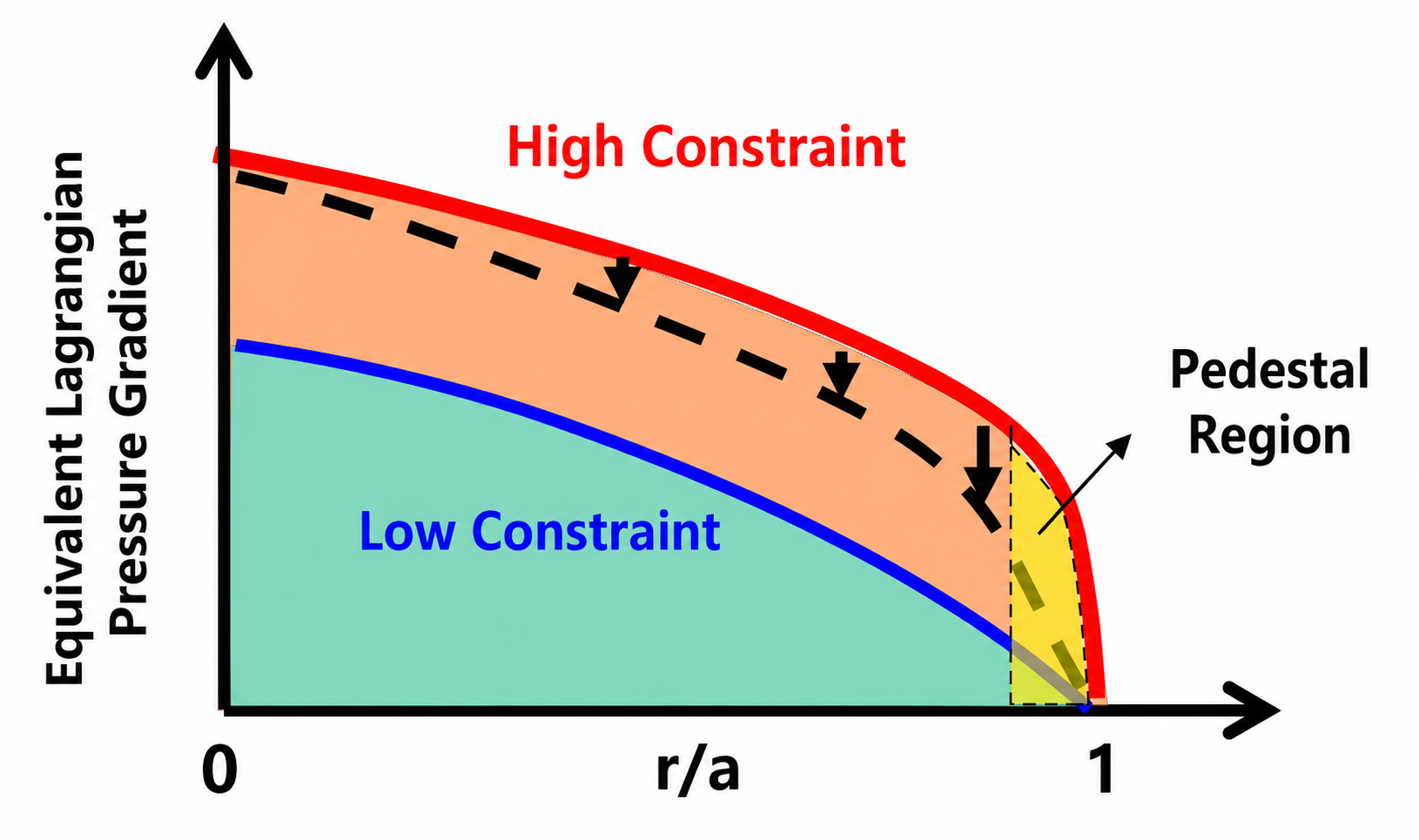}
    \caption{Schematic comparison of the radial plasma profiles in L-mode and H-mode. }
    \label{fig:hmode_pedestal}
\end{figure}

At the heart of magnetic-confinement fusion is the ability to sustain a
sufficiently hot and dense plasma with good energy and particle confinement.
Achieving reactor-relevant performance therefore requires the plasma to operate
in a high-confinement regime. In tokamaks, the transition from low-confinement
mode (L-mode) to high-confinement mode (H-mode) suppresses turbulent transport
near the plasma edge and forms an edge transport barrier. This barrier produces
\textbf{a characteristic pedestal with steep pressure, density, and temperature
gradients}, as illustrated schematically in Fig.~\ref{fig:hmode_pedestal}.
The pedestal plays a central role in determining both the global confinement
performance and the stability limits of the plasma.

\textbf{Within this steep-gradient region, drift-wave instabilities are particularly
important} because they regulate cross-field turbulent transport and thereby
influence the formation and sustainability of the pedestal. \textbf{Understanding
their spatial structure and stability is therefore essential for predicting
the plasma confinement state.} 
In this problem, the steep equilibrium gradients introduce a short radial scale, while
drift-wave perturbations may simultaneously exhibit rapid variation in the
poloidal direction. Consequently, their radial and poloidal structures can no
longer be treated independently. Magnetic shear, characterized by the parameter
$s$, causes the poloidal phase evolution to vary with radial position, whereas
toroidal curvature couples the poloidal geometry to the radial variation of the
perturbation. \textbf{The drift-wave eigenstate to be modeled is therefore a genuinely
two-dimensional mode that is generally nonseparable in the radial and poloidal
coordinates}, rather than a product of independent one-dimensional radial and
poloidal profiles.

Based on the analysis above, the perturbed electrostatic potential is represented by the complex mode field
$\varphi(x,\theta)=\varphi_r(x,\theta)+\mathrm{i}\varphi_i(x,\theta)$.
Its real and imaginary components do not represent separate physical modes. Instead, they jointly describe the spatial amplitude and phase of a single eigenmode. Because an eigenfunction of a homogeneous problem can be multiplied by an arbitrary nonzero complex constant, its overall amplitude and phase remain undetermined unless a normalization condition or nonzero field observations are imposed. The  complex eigenfrequency is
$\omega=\omega_r+\mathrm{i}\omega_i$.
Under the adopted $\mathrm{e}^{-\mathrm{i}\omega t}$ convention, $\omega_r$ determines the oscillation frequency, while positive and negative values of $\omega_i$ correspond to linear growth and damping, respectively. Since $\omega$ enters the coefficients of the governing operators, the spatial mode structure and the eigenfrequency must be determined jointly.

\textbf{The above physical analysis reduces the problem to that of reconstructing a two‑dimensional nonseparable complex mode-field structure and its corresponding complex eigenfrequency from limited observational data.}
For a forward calculation with fixed equilibrium parameters, the governing operators and boundary conditions determine the admissible eigenpairs $(\varphi,\omega)$. The inverse problem considered here provides only sparse samples of the complex mode field, so different candidate fields and eigenfrequencies may agree with the available observations. In addition, the homogeneous governing equation permits the trivial zero field and may admit multiple eigenpairs. Domain-wide collocation residuals, boundary conditions, and nonzero field observations must therefore be combined to constrain the reconstruction toward an eigenpair that is consistent with the adopted physical model. These constraints improve the identifiability of the inverse problem, but do not by themselves establish mathematical uniqueness.

\section{Extended Related Work}

This section reviews forward drift-wave eigenanalysis and discusses two
main challenges in sparse complex-eigenpair inversion: target complex-eigenpair
identification and oscillatory complex mode-field representation.

\subsection{Target Complex-Eigenpair Identification}
\label{sec:target_eigenpair_identification}

Ion-temperature-gradient (ITG) and related drift-wave instabilities are
commonly studied using local ballooning models or global eigenvalue
formulations~\cite{xie2017ballooning,lu2017symmetry}.
Local models resolve the mode along a magnetic field line while approximating
its radial structure through localization assumptions. These approximations
are effective when the equilibrium varies slowly over the radial mode width
and the mode remains localized in the poloidal direction.

In steep-gradient regimes, however, these assumptions become less reliable.
Radial variation, poloidal structure, magnetic shear, and toroidal curvature
are strongly coupled, so the radial and poloidal structures can no longer be
treated independently. The resulting eigenmodes may belong to distinct
branches with different complex eigenfrequencies and may exhibit nonseparable
two-dimensional structures or symmetry breaking associated with local--global
coupling~\cite{lu2017symmetry,zielinski2020global,Qiu2024TwoDimensionalITG}.
Global eigensolvers capture these effects by discretizing the full linear
operator and solving for its complex eigenpairs, provided that the equilibrium,
boundary conditions, and governing operator are known.

The inverse problem considered here differs from this forward setting because
only sparse samples of the mode field are observed. Both the missing field and
the complex eigenfrequency must be inferred while satisfying the prescribed
physical operator. Since the eigenfrequency appears in the operator
coefficients, changing it also changes the spatial balance that defines an
admissible mode. The field and eigenfrequency must therefore be inferred
jointly rather than treated as independent quantities. Physics-informed
learning provides a natural way to combine sparse observations with governing
equations and has been applied to general inverse PDE problems as well as
fusion-plasma transport and equilibrium reconstruction.

Reformulating eigenanalysis as an inverse problem introduces a fundamental
identifiability challenge. Because the eigenvalue equation is homogeneous, the
zero field satisfies it for any candidate eigenfrequency. Moreover, if an
eigenfunction is admissible, multiplying it by any nonzero complex constant
leaves the eigenvalue relation unchanged. The governing equation alone
therefore cannot determine the global amplitude or phase of the mode. Similar
ambiguities in physics-informed eigenvalue solvers are commonly addressed
through normalization, nontriviality, or mode-selection
constraints~\cite{wu2023spectral}.
The existence of multiple eigenmode branches further implies that a small
equation residual does not necessarily identify the desired mode.

Different constraints play complementary roles in resolving these
ambiguities. Sparse nonzero complex observations prevent collapse to the
trivial solution, anchor the field amplitude and phase, and help distinguish
the observed mode from other admissible branches. Boundary conditions restrict
the possible spatial structures, while residuals evaluated at interior
collocation points enforce consistency with the governing operator away from
the observations. Together, these constraints support recovery of the target
nontrivial eigenpair, although they do not guarantee mathematical uniqueness.
Complex-eigenpair inversion must therefore satisfy both physical consistency
and target-mode identification.

\subsection{Oscillatory Complex Mode-Field Representation}

Reliable eigenfrequency recovery also depends on accurately reconstructing the
mode field. Coordinate-based multilayer perceptrons tend to learn smooth,
low-frequency components before rapidly varying ones, a behavior commonly
known as spectral bias~\cite{wang2021eigenvector}. This bias is particularly
restrictive for steep-gradient ITG modes, which may be concentrated within a
narrow radial region while oscillating rapidly in the poloidal direction.

In physics-informed inversion, field approximation errors affect more than the
reconstruction itself. Spatial differentiation can amplify these errors and
distort the governing-equation residual. Because the eigenfrequency is inferred
through the same residual, the optimization may partially compensate for an
inaccurate field by changing the spectral parameter. A small residual may
therefore result from an incorrect balance between the reconstructed field and
the inferred eigenfrequency. This coupling makes the optimization sensitive to
representation errors and imbalances among the different loss
terms~\cite{wang2021gradient}.

Existing techniques address different aspects of this difficulty.
Frequency-aware coordinate encodings improve the representation of
high-frequency and multiscale functions. Domain decomposition and adaptive
sampling improve the resolution of spatially localized structures, while
loss-balancing strategies help reduce optimization imbalance. For a complex
field, separate real-valued output channels provide a basic representation,
but they do not explicitly model real--imaginary interactions during feature
propagation. Complex-valued transformations provide a structured mechanism
for coupling the two components~\cite{trabelsi2018deep,lee2022complex}, yet
they do not by themselves remove spectral bias or prevent compensation between
field and eigenfrequency errors.

The proposed CEI-PINN in this paper combines these complementary mechanisms within a unified
physics-informed framework. Fourier feature encoding represents rapidly
oscillatory structures, complex-valued feature propagation couples the real
and imaginary components, and staged optimization reduces compensation between
mode-field errors and the inferred eigenfrequency. Together, these components
address the coupled representation and identification challenges of sparse
complex-eigenpair inversion.

\section{Reference Dataset and Task Difficulty} \label{sec:supp_dataset}

This section extends the dataset description in the main text
by specifying the provenance and access conditions of the data,
the computational grid, the quantitative characteristics of the
reference mode, and the normalization and sparse-sampling
procedures used in the inverse task.

\subsection{Data Provenance and Controlled Access}

Some key physical parameters used to configure the
physics-based eigensolver were provided by the EAST team at
the Institute of Plasma Physics, Chinese Academy of Sciences
(ASIPP). The reference mode field and complex eigenfrequency
were subsequently generated using the eigensolver.

Because these input parameters and the resulting reference
dataset are subject to the institutional data-management and
confidentiality requirements of ASIPP, the authors are not
authorized to redistribute the complete dataset publicly.
Researchers seeking access may submit a request to the
corresponding author or a designated institutional contact.
Each request will be considered subject to written approval
from ASIPP. Approved access will require an appropriate
data-use and confidentiality agreement specifying the permitted
purpose, scope of use, data-storage requirements, and
redistribution restrictions.

\subsection{Reference Solution and Computational Grid}

A physics-based eigensolver was used to generate a
high-resolution reference solution for the representative ITG
ground-state branch considered in the main text. The reference
dataset contains \(N=321{,}201\) spatial samples arranged on a
tensor-product grid with 801 poloidal locations and 401 radial
locations. The computational coordinates cover
\[
\theta\in[0,2\pi],
\qquad
x\in[0.5975,1.0025].
\]
At each grid point, the dataset stores the complex-valued mode
field
\begin{equation}
    \varphi(x,\theta)
    =
    \varphi_r(x,\theta)
    +
    \mathrm{i}\varphi_i(x,\theta),
\end{equation}
where $\varphi_r$ and $\varphi_i$ denote its real and imaginary
components, respectively. The corresponding reference complex
eigenfrequency is
\begin{equation}
    \omega^{\mathrm{ref}}
    =
    2.0877+2.9867\mathrm{i}.
\end{equation}
All localization and oscillation statistics are computed directly on this grid without interpolation.

\subsection{Quantified Characteristics of the Inverse Task}

The difficulty of the inverse task can be characterized through
three complementary properties of the reference field: strong
spatial localization, rapid poloidal oscillation, and
complex-valued component coupling.

\paragraph{Spatial localization.}
Using $|\varphi_j|^2$ as the discrete field-energy measure at
the $j$th grid point, the highest-energy 8.5\% of the spatial
samples account for approximately 90\% of the total field
energy. After summing the field energy over the poloidal
direction, 90\% of the radially accumulated energy lies within
\[
x\in[0.7494,0.8506],
\]
and the maximum occurs near $x=0.8$. Uniform spatial sampling
therefore allocates most observations to low-amplitude regions
and comparatively few observations to the narrow region
containing most of the mode energy.

\paragraph{Poloidal oscillation.}
Fourier analysis along the poloidal direction identifies a
dominant angular index of 63. At the peak radius $x=0.8$, both
the real and imaginary components exhibit 126 zero crossings
over one complete poloidal period. These diagnostics show that
the target is not a smooth low-frequency field. Sparse
observations must instead resolve a rapidly oscillating
structure whose characteristic variation is concentrated within
a limited radial region.

\paragraph{Complex-valued coupling.}
The real and imaginary components have comparable magnitudes
and jointly determine the local amplitude and phase of the mode
field. Both components enter the complex governing equation and
its spatial derivatives. An error in either component therefore
changes the physical residual and may be partially compensated
by a change in the estimated complex eigenfrequency. The inverse
task consequently requires the two field components and the
global spectral parameter to be recovered in a mutually
consistent manner.

\subsection{Field Normalization and Sparse Supervision}

Before the supervised and evaluation locations are separated,
the complete reference field is normalized using a single
positive real scaling factor. Let $\varphi_j$ denote the complex
field value at the $j$th grid point. The shared max-absolute
scaling factor is defined as
\begin{equation}
    s_{\varphi}
    =
    \left[
    \max_{1\leq j\leq N}
    \left\{
    \left|\operatorname{Re}(\varphi_j)\right|,
    \left|\operatorname{Im}(\varphi_j)\right|
    \right\}
    \right]^{-1},
\end{equation}
and the normalized field is
\begin{equation}
    \widetilde{\varphi}_j
    =
    s_{\varphi}\varphi_j.
\end{equation}
Applying the same positive scaling factor to both components
preserves their relative magnitudes and phase relationship. No
independent component-wise normalization is performed.

After normalization, the supervised locations are sampled
uniformly without replacement from the complete spatial grid
using a fixed random seed. At the default supervision ratio of
6\%, the supervised set contains\(N_d=19{,}272\)
spatial locations selected from the full set of 321,201 grid
points. Both the real and imaginary field values are supplied at
every selected location, so each supervised sample retains the
local complex-field information.

The field values at the remaining locations are not used as
supervised targets and are retained for full-field evaluation.
This sampling procedure applies only to the supervised field
observations. Boundary samples and interior collocation points
used to evaluate the governing-equation residual form separate
constraint sets.

\section{PINN Formulation and Coupled Inversion}

\subsection{Conventional Eigensolvers and Sparse Inversion}

Physics-informed neural networks provide an alternative inverse-learning framework by incorporating governing-equation residuals, boundary constraints, and observation errors into a unified training objective. This formulation makes it possible to use sparse observations while maintaining consistency with the underlying physical model. However, its direct application to ITG complex eigenfrequency inversion remains challenging because the complex mode field and complex eigenfrequency must be recovered simultaneously.

\subsection{Limitations of Standard PINNs for ITG Inversion}
\label{sec:supp_pinn_limitations}

The main text summarizes the principal difficulties encountered
by standard PINNs in ITG inversion. This section further explains
how these difficulties arise from the quantified properties of
the reference field and how they affect eigenpair recovery.

\textbf{(1) Representation under sparse sampling.}
The reference field is strongly concentrated within a narrow
radial region, whereas the supervised locations are sampled
uniformly over the complete computational domain. Consequently,
most observations are located outside the region containing the
dominant mode energy. Within the localized region, the field also
varies rapidly along the poloidal direction. At the peak radius,
the dominant angular index is 63, and both field components
exhibit 126 zero crossings over one poloidal period. A standard
PINN must therefore infer many oscillation cycles from relatively
few observations in the most informative region.Moreover, the smooth approximations favored by standard PINNs may reproduce the overall envelope and match isolated observations, yet still miss complete oscillation cycles or introduce local phase shifts between neighboring samples.Such errors may appear small in the pointwise field loss but become much more pronounced after spatial differentiation. Because the governing operator contains both radial and poloidal derivatives, errors in the local wavelength,
phase, or peak position directly contaminate the physics residual. Accurately fitting sparse field observations is therefore insufficient unless the oscillatory structure between them is also reconstructed.

\textbf{(2) Interaction between the field components.}
The real and imaginary components have comparable amplitudes and
together determine the local amplitude and phase of the ITG mode.
Their relationship is spatially varying rather than being
described by a constant phase offset. Moreover, the real and
imaginary parts of the governing equation both depend on the two
field components, their derivatives, and the real and imaginary
parts of the eigenfrequency. An error in either field component
therefore contributes to both parts of the physical residual.

When the two components are treated merely as real-valued output
channels, their interaction is imposed mainly through the final
loss function. The hidden features are not explicitly required to
preserve their relative amplitude and phase during propagation.
As a result, the network may obtain small component-wise data
errors at the supervised locations while producing an
inconsistent complex phase in the unsupervised region. This
inconsistency subsequently enters the spatial derivatives and
disturbs the eigenfrequency inferred from the physical residual.

\textbf{(3) Compensation during joint inversion.}
The mode field contains a large number of spatial degrees of
freedom, whereas the complex eigenfrequency consists of only two
global unknowns. At every collocation point, however, both sets of
variables enter the same governing-equation residual. The
optimizer can consequently reduce the residual either by
correcting the mode structure or by changing the eigenfrequency.
Because sparse observations constrain the field only at selected
locations, the network retains considerable freedom to modify the
field elsewhere in response to an inaccurate eigenfrequency.

This interaction is particularly important during early training,
when the oscillatory mode structure has not yet been accurately
reconstructed. An error in the provisional field may drive the
eigenfrequency toward a compensatory value. Conversely, once the
eigenfrequency moves away from the target value, the field network
may adapt its unsupervised predictions to remain compatible with
that value. The total training loss can therefore decrease even
though the recovered field and eigenfrequency do not form the
correct eigenpair.

These effects reinforce one another in standard PINN inversion.
Insufficient representation of the localized oscillations produces
field and derivative errors; these errors perturb the physical
residual and bias the eigenfrequency update; the biased
eigenfrequency then further alters the reconstructed mode field.
This feedback explains why a standard jointly trained PINN may
converge to a low-loss solution without accurately recovering the
target ITG eigenpair.

\subsection{Challenge-Driven Framework Design} \label{sec:supp_challenge_driven_design}

The architecture and training procedure are presented in the main
text. This section further explains how each design component
modifies the representation or optimization process in response
to the corresponding inversion challenge.

\textbf{(1) Role of Fourier feature encoding.}
The normalized spatial coordinate is retained together with paired
sine and cosine features at multiple frequencies. These inputs
serve complementary purposes. The original coordinate provides
information about the slowly varying envelope and radial
localization of the mode, whereas the sinusoidal features provide
direct access to its rapid poloidal variations. Each sine--cosine
pair forms a phase-complete representation at the corresponding
frequency, allowing oscillations with different local phases to be
constructed through combinations of the encoded features.

This mapping exposes the required frequency content before the
coordinates enter the nonlinear hidden layers. The network
therefore does not need to generate all high-frequency components
progressively from a low-dimensional coordinate input. Because the
mapping is deterministic and differentiable, spatial derivatives
of the predicted field can still be evaluated with respect to the
original coordinates through automatic differentiation. The
physical equation is consequently enforced on the predicted field
without spatial interpolation or preprocessing of its derivatives.

\textbf{(2) Complex-valued feature propagation.}
The Fourier-encoded real feature vector $\boldsymbol{z}$ is first
duplicated and stacked as
\begin{equation}
    \boldsymbol{q}^{(0)}
    =
    \begin{bmatrix}
        \boldsymbol{z} \\
        \boldsymbol{z}
    \end{bmatrix}.
\end{equation}
Each hidden layer then applies the block-structured transformation
\begin{equation}
    \boldsymbol{q}^{(\ell+1)}
    =
    \tanh\left(
    \begin{bmatrix}
        \boldsymbol{A}^{(\ell)}
        &
        \boldsymbol{B}^{(\ell)}
        \\
        -\boldsymbol{B}^{(\ell)}
        &
        \boldsymbol{A}^{(\ell)}
    \end{bmatrix}
    \boldsymbol{q}^{(\ell)}
    +
    \begin{bmatrix}
        \boldsymbol{c}^{(\ell)} \\
        \boldsymbol{d}^{(\ell)}
    \end{bmatrix}
    \right),
\end{equation}
where $\tanh(\cdot)$ is applied component-wise. The off-diagonal
blocks allow each feature group to depend on the other throughout
the network, thereby maintaining interaction between the real and
imaginary representations.

The output layer uses the same block structure without an
activation function and divides the resulting vector into
\begin{equation}
    \widehat{\varphi}
    =
    \widehat{\varphi}_r
    +
    \mathrm{i}\widehat{\varphi}_i.
\end{equation}
The two outputs are subsequently used for automatic differentiation
and complex physical-residual evaluation.

\textbf{(3) Role of the three-stage training strategy.}
The staged schedule controls which variables are allowed to respond
to the physical residual at different points in training. During
the initial field-fitting stage, the sparse observations and
boundary constraints first establish a nontrivial approximation of
the mode amplitude, phase, and localization. Introducing the
physical residual only after this initial anchoring prevents an
inaccurate early residual from dominating the formation of the
field structure.

During eigenfrequency identification, the field network is frozen.
The optimizer can no longer reduce the residual by freely modifying
the high-dimensional mode structure and must instead correct the
two components of the complex eigenfrequency. This converts the
central identification step into a lower-dimensional optimization
problem based on an observation-consistent provisional field.

After the eigenfrequency has been identified, fixing it during
field refinement prevents subsequent changes in the mode structure
from shifting the recovered spectral parameter. The remaining
optimization can then improve agreement with the observations,
boundary conditions, and governing equation under a fixed global
eigenfrequency. The staged procedure therefore changes the
conditioning of the inverse problem rather than merely dividing a
joint optimization into several training intervals.

The three components operate as a connected design. Fourier
features provide an initial representation capable of resolving
the oscillatory field, complex-valued propagation maintains
real--imaginary interaction within that representation, and staged
training prevents the remaining field errors from being absorbed
into the eigenfrequency. Their combination links representation
accuracy directly to stable complex-eigenpair identification under
sparse supervision.

\subsection{Standard Complex Eigenfrequency Inversion}

Consider the simplified two-dimensional complex eigenvalue
problem
\begin{equation}
    \frac{\partial^2\phi}{\partial x^2}
    +
    \frac{\partial^2\phi}{\partial\theta^2}
    +
    \omega\phi
    =
    0,
    \label{eq:supp_standard_model}
\end{equation}
where $\phi(x,\theta)\in\mathbb{C}$ denotes the complex mode
field and
\begin{equation}
    \omega
    =
    \omega_r+\mathrm{i}\omega_i
\end{equation}
denotes the unknown complex eigenfrequency. A standard PINN
approximates the mode field as
\begin{equation}
    \widehat{\phi}(x,\theta)
    =
    f_{\vartheta}(x,\theta),
\end{equation}
where $\vartheta$ denotes the network parameters. The
eigenfrequency components $\omega_r$ and $\omega_i$ are treated
as additional trainable parameters.

Substituting the predicted field into
Eq.~\eqref{eq:supp_standard_model} gives the physical residual
\begin{equation}
    r(x,\theta)
    =
    \frac{\partial^2\widehat{\phi}}{\partial x^2}
    +
    \frac{\partial^2\widehat{\phi}}{\partial\theta^2}
    +
    \omega\widehat{\phi}.
    \label{eq:supp_physical_residual}
\end{equation}
For $N_f$ interior collocation points
$\{(x_j^f,\theta_j^f)\}_{j=1}^{N_f}$, the PDE loss is defined as
\begin{equation}
    \mathcal{L}_{\mathrm{pde}}
    =
    \frac{1}{N_f}
    \sum_{j=1}^{N_f}
    \left|
        r(x_j^f,\theta_j^f)
    \right|^2.
    \label{eq:supp_pde_loss}
\end{equation}
This term measures violations of the governing equation at the
sampled interior points and directly couples
$\widehat{\phi}$ to the trainable complex eigenfrequency
$\omega$.

For any complex scalar $z$, the squared modulus used in all
loss terms is
\begin{equation}
    |z|^2
    =
    zz^{*}
    =
    \left[\operatorname{Re}(z)\right]^2
    +
    \left[\operatorname{Im}(z)\right]^2,
\end{equation}
where $(\cdot)^{*}$ denotes complex conjugation.

The predicted field is also required to satisfy the prescribed
boundary condition
\begin{equation}
    \widehat{\phi}(x,\theta)
    =
    g(x,\theta),
    \qquad
    (x,\theta)\in\partial\Omega,
\end{equation}
where $g(x,\theta)$ denotes the boundary value. For $N_b$
boundary samples
$\{(x_j^b,\theta_j^b)\}_{j=1}^{N_b}$, the boundary loss is
\begin{equation}
    \mathcal{L}_{\mathrm{bc}}
    =
    \frac{1}{N_b}
    \sum_{j=1}^{N_b}
    \left|
        \widehat{\phi}(x_j^b,\theta_j^b)
        -
        g(x_j^b,\theta_j^b)
    \right|^2.
    \label{eq:supp_boundary_loss}
\end{equation}

Given $N_d$ sparse complex-field observations
$\{(x_j^d,\theta_j^d,\phi_j^{*})\}_{j=1}^{N_d}$, the data loss
is
\begin{equation}
    \mathcal{L}_{\mathrm{data}}
    =
    \frac{1}{N_d}
    \sum_{j=1}^{N_d}
    \left|
        \widehat{\phi}(x_j^d,\theta_j^d)
        -
        \phi_j^{*}
    \right|^2.
    \label{eq:supp_data_loss}
\end{equation}
The data loss anchors the predicted complex field to the
available observations at the sparsely supervised locations.

The three constraints are combined into the total objective
\begin{equation}
    \mathcal{L}_{\mathrm{total}}
    =
    \lambda_{\mathrm{pde}}
    \mathcal{L}_{\mathrm{pde}}
    +
    \lambda_{\mathrm{bc}}
    \mathcal{L}_{\mathrm{bc}}
    +
    \lambda_{\mathrm{data}}
    \mathcal{L}_{\mathrm{data}}.
    \label{eq:supp_total_loss}
\end{equation}
The weights $\lambda_{\mathrm{pde}}$,
$\lambda_{\mathrm{bc}}$, and $\lambda_{\mathrm{data}}$ balance
the governing-equation, boundary, and sparse-data constraints,
respectively. Direct minimization of
$\mathcal{L}_{\mathrm{total}}$ jointly updates the network
parameters $\vartheta$ and the eigenfrequency components
$\omega_r$ and $\omega_i$.

For the complete ITG problem, the network prediction is
substituted into the global governing equation, and the physical
residual is evaluated from
\begin{equation}
    (L_0+L_1)\widehat{\phi}(x,\theta).
\end{equation}
The definitions of $\mathcal{L}_{\mathrm{data}}$,
$\mathcal{L}_{\mathrm{bc}}$, and
$\mathcal{L}_{\mathrm{pde}}$ retain the same roles, while the
simplified residual in Eq.~\eqref{eq:supp_physical_residual} is
replaced by the complete ITG residual. The physical parameters
appearing in $L_0$ and $L_1$ are summarized in
Table~\ref{tab:physical_parameters}.

\begin{table}[t]
\centering
\small
\caption{Definitions of physical parameters.}
\label{tab:physical_parameters}
\renewcommand{\arraystretch}{1.15}
\setlength{\tabcolsep}{4pt}
\begin{tabular}{c|l}
\hline
Parameter & Description \\
\hline

$\rho_i$
& Ion gyroradius \\

$k_\theta=nq(r_0)/r_0$
& Poloidal wavenumber \\

$n$
& Toroidal mode number \\

$q$
& Safety factor \\

$s=\mathrm{d}\ln q/\mathrm{d}\ln r$
& Magnetic shear \\

$\omega_{*e}=k_\theta T_e/(eBL_n)$
& Electron diamagnetic frequency \\

$T_e$
& Electron temperature \\

$e$
& Elementary charge \\

$B$
& Magnetic-field strength \\

$L_n$
& Density-gradient scale length \\

$\epsilon_n=L_n/R$
& Normalized density-gradient parameter \\

$R$
& Major radius \\

$\eta_i=L_n/L_{T_i}$
& Ion temperature-gradient parameter \\

$L_{T_i}$
& Ion-temperature-gradient scale length \\

$\tau=T_e/T_i$
& Electron-to-ion temperature ratio \\

$T_i$
& Ion temperature \\

$\delta_e$
& Trapped-electron collisionality parameter \\
\hline
\end{tabular}
\end{table}

\subsection{Coupled Inversion Challenges}
\label{sec:supp_pinn_limitations}

The limitations of standard PINN inversion are coupled through
the physical residual. Errors in the mode-field representation
affect its spatial derivatives, errors in the real or imaginary
component alter the complex residual, and the resulting residual
error can influence the estimated eigenfrequency.

\paragraph{Representation of localized oscillatory fields.}

Under steep-gradient conditions, the ITG mode field is strongly
localized in the radial direction and rapidly oscillatory in
the poloidal direction. Standard multilayer perceptrons
typically learn smooth, low-frequency components before
high-frequency components. Consequently,
$\widehat{\phi}$ may reproduce the broad spatial envelope while
failing to resolve the local oscillatory structure.

This discrepancy becomes more serious after spatial
differentiation. A relatively small error in
$\widehat{\phi}$ may produce a substantially larger error in its
first- or second-order derivatives. The resulting derivative
errors distort $r(x,\theta)$ and reduce the reliability of
$\mathcal{L}_{\mathrm{pde}}$ as a constraint on the complex
eigenfrequency.

\paragraph{Interaction between real and imaginary components.}

The predicted complex field can be written as
\begin{equation}
    \widehat{\phi}(x,\theta)
    =
    \widehat{\phi}_{r}(x,\theta)
    +
    \mathrm{i}\widehat{\phi}_{i}(x,\theta).
\end{equation}
Using two real-valued outputs is mathematically sufficient to
represent this field. However, if the corresponding features
are processed independently, their interaction is imposed only
indirectly through the loss functions and the governing
equation.

The real and imaginary components jointly determine the local
amplitude and phase of the mode field. An error in either
component therefore changes both the predicted complex
structure and the physical residual. Independent feature
processing may consequently make it more difficult to maintain
consistent spatial structures and phase relationships during
optimization.

\paragraph{Compensation between the mode field and eigenfrequency.}

The predicted field $\widehat{\phi}$ and the complex
eigenfrequency $\omega$ are coupled through $r(x,\theta)$.
When $\vartheta$, $\omega_r$, and $\omega_i$ are optimized
simultaneously, an error in the predicted field may be partially
compensated by a change in the eigenfrequency. Conversely, an
inaccurate eigenfrequency may be accommodated by modifying the
predicted spatial structure.

This compensation can reduce $\mathcal{L}_{\mathrm{pde}}$
without correcting the variable that originally caused the
residual error. It may also produce poorly conditioned or
strongly unbalanced gradients and increase sensitivity to
initialization, residual scaling, and loss weighting. The
homogeneous zero solution, the complex scaling freedom of the
eigenfunction, and the possible presence of multiple eigenmode
branches further enlarge the set of candidate solutions.

These mechanisms form a coupled error pathway. An inaccurate
representation of the oscillatory field produces derivative
errors, the derivative errors distort the physical residual,
and the distorted residual can bias the inferred complex
eigenfrequency. Standard joint optimization does not explicitly
separate these effects.

\subsection{From Inversion Challenges to CEI-PINN Design}
\label{sec:supp_challenge_driven_design}

The preceding analysis identifies three corresponding design
requirements. The network must resolve the spatial frequencies
of the localized ITG mode, maintain interaction between the real
and imaginary features, and limit compensation between the mode
field and eigenfrequency during optimization.

Fourier feature encoding addresses the first requirement by
mapping the spatial coordinates into a representation containing
multiple frequency scales. Complex-valued feature propagation
addresses the second requirement by allowing the real and
imaginary features to interact throughout the network. The two
components therefore serve different functions: Fourier
features improve spatial-frequency representation, whereas
complex-valued propagation strengthens component interaction.

The three-stage training strategy addresses the third
requirement. The procedure first establishes the mode-field
representation, then optimizes $\omega_r$ and $\omega_i$ with
the field network fixed, and finally refines the field network
with the identified eigenfrequency $\widehat{\omega}$ fixed.
This schedule does not treat the mode field and eigenfrequency
as physically independent. Instead, it controls which variables
can respond to the physical residual at each stage and thereby
reduces parameter compensation during optimization.

The three designs act at different points in the coupled error
pathway. Fourier feature encoding targets oscillatory-field
representation, complex-valued feature propagation targets
real--imaginary interaction, and staged training targets
field--eigenfrequency compensation. The governing equation,
boundary conditions, and sparse observations remain unchanged.

\section{Experimental Details and Analyses}
\label{sec:supp_experiments}

\subsection{Network Setup and Preprocessing}
\label{sec:supp_implementation}

CEI-PINN takes the radial coordinate $x$, poloidal coordinate $\theta$, and control parameter $k_{\theta}$ as inputs. Before entering the network, $x$ and $\theta$ are normalized to $[-1,1]$ and transformed using the Fourier feature mapping defined in the main text. The control parameter $k_{\theta}$ is scaled separately and concatenated with the encoded spatial features. Eight Fourier frequencies are used for each normalized spatial coordinate, producing $1+2\times 8=17$ features per coordinate. After concatenating the encoded features of $x$ and $\theta$ with the scaled $k_{\theta}$, the total input dimension is $2(1+2\times 8)+1=35$. The complex-valued network uses the architecture $[35,128,128,128,1]$, comprising three hidden layers of width 128 with hyperbolic tangent activation functions. Its two output channels represent the real and imaginary components of the complex mode field. The real and imaginary parts of the complex eigenfrequency are implemented as two independent real-valued trainable parameters rather than outputs of the field network. The network and eigenfrequency parameters are optimized according to the three-stage training strategy described in the main text.

\subsection{Training Configuration}
\label{sec:supp_training_configuration}

\begin{figure}[t]
\centering
\includegraphics[width=\linewidth]{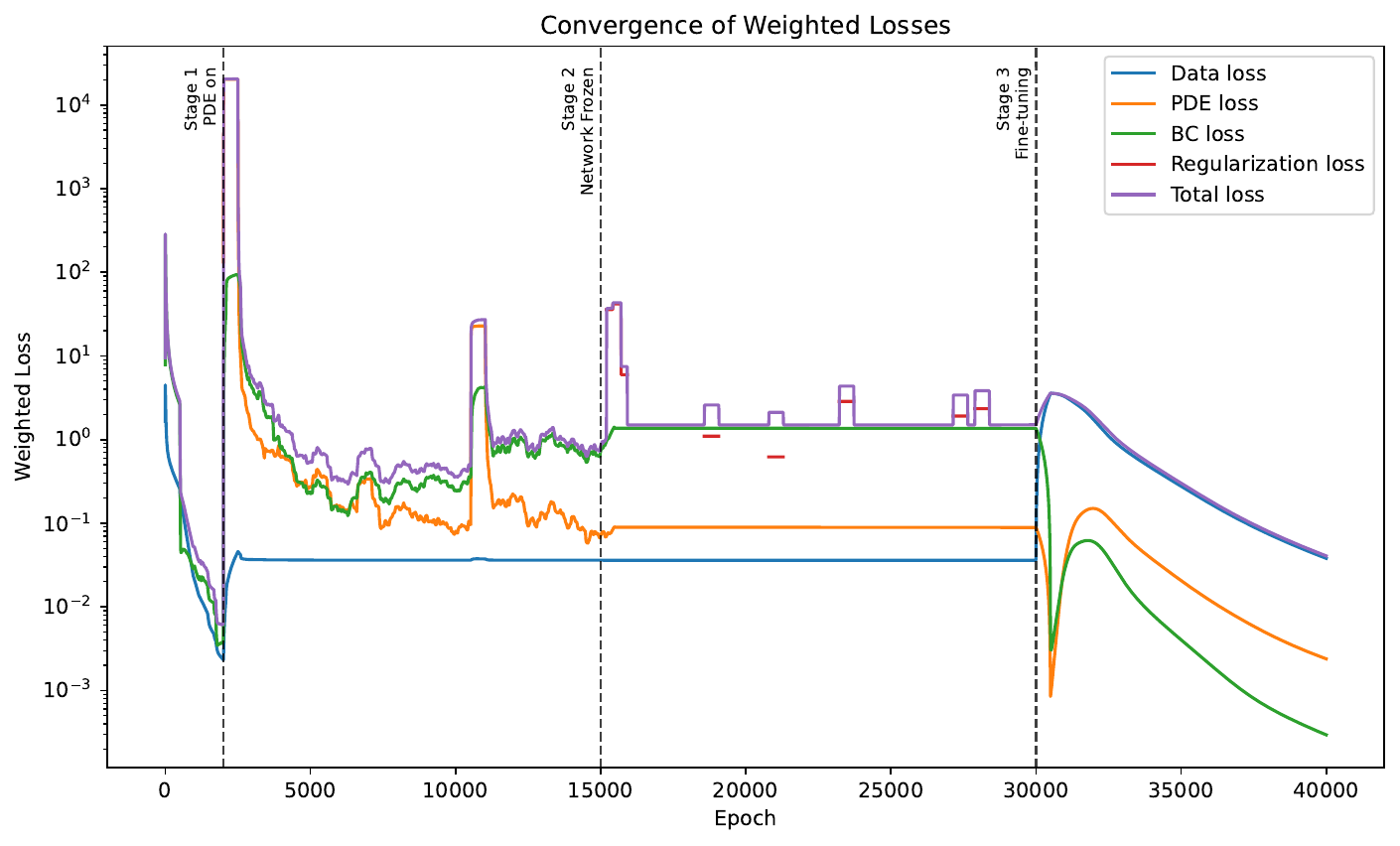}
\caption{Weighted-loss convergence in three-stage training. Dashed lines at epochs $2{,}000$, $15{,}000$, and $30{,}000$ mark PDE-loss start, eigenfrequency fitting with the field net frozen, and field tuning with fixed $\hat{\omega}$, respectively.
}
\label{fig:supp_training_loss}
\end{figure}

All compared models were evaluated using the same sparse observations, boundary samples, PDE collocation points, and evaluation grid. The same supervision ratio and physical constraints were used to ensure a consistent comparison. As shown in Fig.~\ref{fig:supp_training_loss}, training was conducted in three stages, with the weighted losses evolving according to the activation and parameter-freezing schedule described below. During Stage I, the mode-field network was initialized using the data and boundary losses, and the PDE residual loss was activated after the initial field-fitting period. Stage I continued until epoch 15,000. During Stage II, from epochs 15,000 to 30,000, the mode-field network was frozen and only the complex eigenfrequency parameters $\omega_r$ and $\omega_i$ were optimized. During Stage III, from epochs 30,000 to 40,000, the identified eigenfrequency $\hat{\omega}$ was fixed and the mode-field network was refined under the data, boundary, and PDE constraints. This configuration separates mode-field initialization, complex eigenfrequency identification, and fixed-eigenfrequency field refinement, thereby reducing parameter compensation during coupled inversion. 

\subsection{Ablation of Loss Components}
\label{sec:supp_loss_ablation}

To quantify the roles of the individual loss terms, we separately
removed the data-supervision loss, boundary-condition loss, PDE
residual loss, and eigenfrequency regularization term. The results
are reported in Table~\ref{tab:loss_ablation}.

The full model converged to $\widehat{\omega}=2.175+2.927\mathrm{i}$. Without data supervision, the model converged to $\widehat{\omega}=2.168+3.493\mathrm{i}$, indicating that the physical and boundary constraints alone were insufficient to reliably select the target mode. Removing the boundary loss produced a biased solution of $\widehat{\omega}=2.014+3.495\mathrm{i}$, showing that the boundary constraint restricts the feasible solution space and stabilizes the inversion.

Without the PDE residual loss, the eigenfrequency remained near its initial value of $\omega^{(0)}=2.00+3.50\mathrm{i}$, yielding $\widehat{\omega}\approx\omega^{(0)}$. This result confirms that the governing equation provides the central coupling between the predicted mode field and complex eigenfrequency. Finally, removing the regularization term caused the solution to collapse to $\widehat{\omega}=1.021+0.045\mathrm{i}$, indicating that this term prevents convergence to a degenerate nonphysical solution.

\begin{table}[t]
\centering
\caption{Loss-component ablation for complex eigenfrequency inversion.}
\label{tab:loss_ablation}
\begin{tabular}{lcc}
\toprule
\textbf{Ablation}
& \textbf{Converged {$\omega$}}
& \textbf{Status} \\
\midrule
Baseline
& $2.175+2.927\mathrm{i}$
& Converged \\
No-Data
& $2.168+3.493\mathrm{i}$
& Converged \\
No-BC
& $2.014+3.495\mathrm{i}$
& Biased \\
No-PDE
& $2.00+3.50\mathrm{i}$
& Stalled at init \\
No-Reg
& $1.021+0.045\mathrm{i}$
& Collapsed \\
\bottomrule
\end{tabular}
\end{table}

\subsection{Visualization of Initialization Robustness}
\label{sec:supp_initialization_robustness}

\begin{figure}[h!]
\centering
\includegraphics[width=\columnwidth]{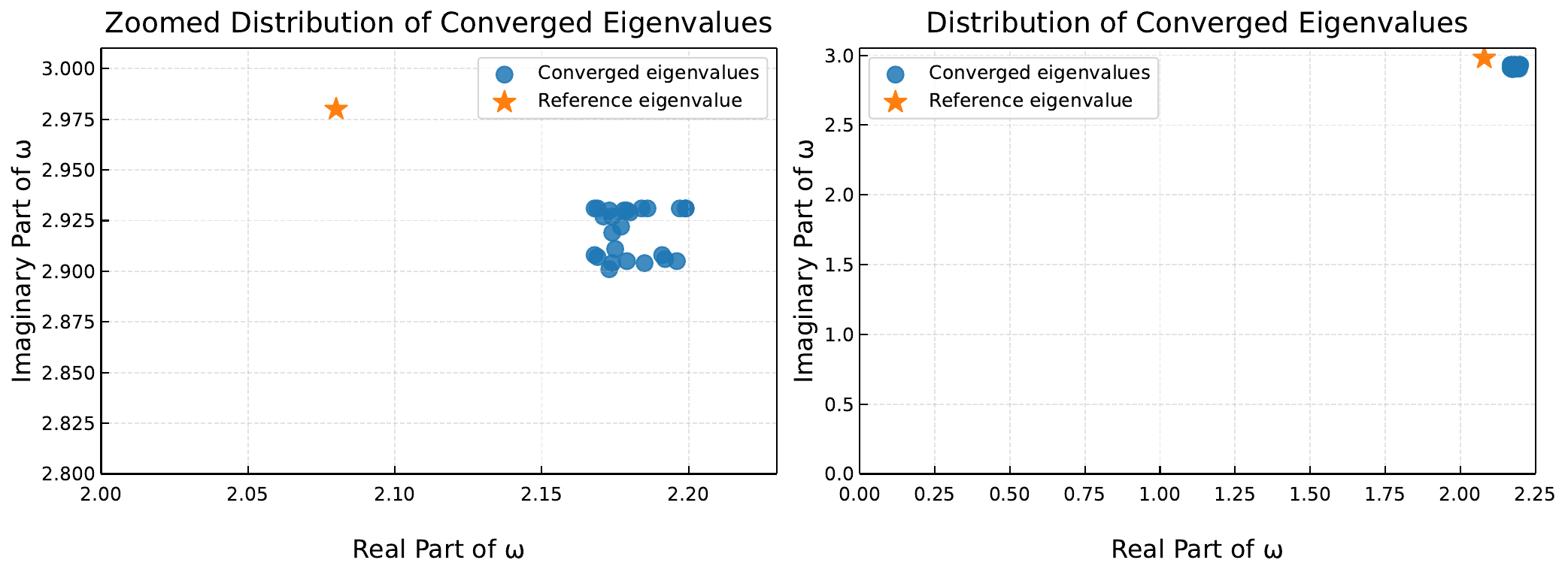}
\caption{Distribution of converged eigenfrequencies across initializations. The left panel shows an enlarged view near the reference, and the right panel shows the full distribution.}
\label{fig:eigenvalue_distribution}
\end{figure}

Figure~\ref{fig:eigenvalue_distribution} visualizes the converged
complex eigenfrequencies obtained from different initial values of $\omega$.
The converged estimates cluster near the reference eigenfrequency across the
tested initialization range, supporting the robustness indicated by the
quantitative results in the main text.

\subsection{Efficiency Comparison}
\label{sec:supp_efficiency}

Table~\ref{tab:efficiency} compares the trained PINN and the
conventional physical solver on the same $801\times401$ reference
grid containing $321{,}201$ spatial points. The runtime of the
physical solver includes matrix construction, solution of the
generalized eigenvalue problem, and reconstruction of the complete
mode field. The reported PINN inference time corresponds to one
forward evaluation after training and excludes the training time.

From Table~\ref{tab:efficiency}, the PINN required $0.0043\,\mathrm{s}$ per evaluation, compared
with $3.678\,\mathrm{s}$ for the physical solver, providing an
approximately $855.3$-fold speedup after training.

\begin{table}[h]
\centering
\caption{Efficiency comparison between the physical solver and PINN.}
\label{tab:efficiency}
\setlength{\tabcolsep}{8pt}
\renewcommand{\arraystretch}{1.12}
\begin{tabular}{lcc}
\toprule
\textbf{Method} & \textbf{Training Time} & \textbf{Inference Time} \\
\midrule
Physical Solver & -- & $3.678\,\mathrm{s}$ \\
PINN & $47\,\mathrm{min}\ 30\,\mathrm{s}$ & $0.0043\,\mathrm{s}$ \\
\bottomrule
\end{tabular}
\end{table}

\end{document}